\documentclass{article}

\usepackage{arxiv}

\usepackage{amssymb}  
\usepackage{amsthm}
\usepackage{amsmath}
\usepackage[utf8]{inputenc} 
\usepackage[T1]{fontenc}    
\usepackage{hyperref}       
\usepackage{cleveref}
\usepackage{url}            
\usepackage{booktabs}       
\usepackage{amsfonts}       
\usepackage{nicefrac}       
\usepackage{microtype}      
\usepackage{lipsum}
\usepackage{parskip}
\usepackage{graphicx}
\usepackage{float}
\usepackage{xcolor}
\usepackage[normalem]{ulem}
\graphicspath{{plots/}}
\usepackage{caption}
\usepackage[labelformat=empty, position=top]{subcaption}
\usepackage[export]{adjustbox}
\usepackage{bm}
\usepackage[round]{natbib}
\usepackage[switch]{lineno}

\title{A deep mixture density network for outlier-corrected interpolation of crowd-sourced weather data}

\author{
  Charlie Kirkwood\\
  Department of Mathematics\\
  University of Exeter, UK\\
  \texttt{c.kirkwood@exeter.ac.uk} \\
  \And
  Theo Economou\\
  Department of Mathematics\\
  University of Exeter, UK\\
  \&\\
  Climate and Atmosphere Research Centre\\
  The Cyprus Institute, Cyprus\\
  \And
  Nicolas Pugeault\\
  School of Computing Science\\
  University of Glasgow, UK\\
  \And
  Henry Odbert\\
  Met Office Applied Science\\
  Met Office, UK\\
}

\begin{document}
\maketitle

\begin{abstract}

As the costs of sensors and associated IT infrastructure decreases --- as exemplified by the Internet of Things --- increasing volumes of observational data are becoming available for use by environmental scientists. However, as the number of available observation sites increases, so too does the opportunity for data quality issues to emerge, particularly given that many of these sensors do not have the benefit of official maintenance teams. To realise the value of crowd sourced `Internet of Things' type observations for environmental modelling, we require approaches that can automate the detection of outliers during the data modelling process so that they do not contaminate the true distribution of the phenomena of interest. To this end, here we present a Bayesian deep learning approach for spatio-temporal modelling of environmental variables with automatic outlier detection. Our approach implements a Gaussian-uniform mixture density network whose dual purposes --- modelling the phenomenon of interest, and learning to classify and ignore outliers --- are achieved simultaneously, each by specifically designed branches of our neural network. For our example application, we use the Met Office's Weather Observation Website data, an archive of observations from around 1900 privately run and unofficial weather stations across the British Isles. Using data on surface air temperature, we demonstrate how our deep mixture model approach enables the modelling of a highly skilled spatio-temporal temperature distribution without contamination from spurious observations. We hope that adoption of our approach will help unlock the potential of incorporating a wider range of observation sources, including from crowd sourcing, into future environmental models.

\end{abstract}


\section*{Introduction}

As environmental scientists, the volumes of observational data that we have at our disposal are ever increasing. Movements such as the Internet-of-Things (IoT), exemplified in the context of weather data by the Met Office's Weather Observation Website \citep{kirk2021weather}, have enabled near real-time collection and sharing of environmental data by low cost sensing equipment around the world. The implications of this are numerous, but where once the challenge was to collect sufficient data for specific modelling problems (often hindered by expense), now often the challenge is to maximise the utility of the high volumes of data that we already have. The rise of `data science' can be viewed, to some extent, as a response to this shift in challenges. 

In the case of weather modelling and forecasting, it is likely that harnessing the ever-growing network of IoT type environmental sensor data, in addition to the observations provided by traditional official weather stations, will facilitate the development of higher-precision, finer-scale models which can serve more specific predictions to stakeholders \citep{bell2015good, chapman2017can}. Linked to this, a benefit of IoT type sensor data is that these observations have the potential to be more representative of the weather experienced by the device owners themselves (e.g. due to private weather stations being located at homes), rather than representative of remote rural locations (as tends to be the case for official weather stations). The adoption of data from these unofficial private weather stations and IoT type environmental sensors could therefore enable models to provide more specific, personalised weather information at hyperlocal scales.

However, while crowd-sourcing weather data can greatly increase the number of observations being made, and the number of unique locations which are observed, it also opens the door to data quality issues owing to the low cost, low maintenance nature of unofficial weather stations compared to official weather stations. The traditional way of addressing data quality issues is to have some form of manually-guided rules based quality control procedure to subjectively approve or deny the inclusion of each sensor's observations into downstream models. While a manually-guided procedure may seem the best approach in terms of having complete hands-on control at an individual observation level, such an approach will tend to suffer from scalability issues as the number of sensors increases, and is difficult to achieve consistently through space and time.

As the number of sensors enters or exceeds the thousands, it becomes necessary to automate aspects of the quality control procedure in order to keep up with the scale of the task. Common approaches include statistical time-series analysis or rule-based outlier detection algorithms to help identify sensors that are producing data of questionable quality, which can then be excluded from input into subsequent models. Here we propose a unified approach whereby detection of outliers is achieved as part of a downstream statistical data model itself: in this case a Bayesian deep neural network based spatio-temporal interpolator of crowd-sourced temperature observations collected by the Met Office's Weather Observation Website, with a mixture model or mixture density network architecture to enable automatic identification and correction of outliers as part of the modelling process.

In this paper we proceed by briefly providing some background on IoT sensor data and its potential benefits for environmental modelling applications, as well as an overview of existing methods for outlier detection. We then introduce our deep mixture model approach for spatio-temporal interpolation with simultaneous probabilistic outlier detection, using an example dataset composed of surface temperature observations collected by the Met Office's Weather Observation Website. By adopting a mixture model approach, we incorporate our knowledge about data issues into the data model through our choice of probability distributions, which provide our likelihood function. This, in combination with our Bayesian approach allows us to quantify both aleatoric and epistemic uncertainties --- uncertainty about the data and uncertainty about the fit of the model --- in order to provide a well-calibrated posterior predictive distribution. Bayesian deep learning frameworks (here we use Tensorflow Probability) allow us to combine the above benefits of Bayesian statistical modelling with the flexibility and scalability of deep neural networks.

We assess the performance of our model on held-out test data, finding our approach to be successful in filtering outliers in order to provide `clean' spatio-temporal interpolation that is free from outlier-induced anomalies. In addition, as our probabilistic approach (including epistemic uncertainty via Monte Carlo dropout as a Bayesian approximation) provides a well-calibrated predictive distribution rather than single point predictions, it therefore provides useful information for downstream applications and decision making.

\section*{Background}
\subsection*{IoT sensor data in environmental modelling}

Within the field of  environmental modelling, the concept of `models of everywhere' \citep{beven2007towards} has been proposed. This is a concept which stems specifically from hydrology but is applicable across environmental sciences. The concept aims to ``change the nature of the modelling process, from one in which general model structures are used in particular catchment applications to one in which modelling becomes a learning process about places" \citep{beven2012modelling}. This idea is driven by the need to  constrain uncertainty in the modelling process in order to support policy setting and decision making \citep{blair2019models}. The concept is a reaction to the shortfalls of the use of `generic models', in which spatially-discretised (gridded) predictions are likely to fail to provide well-calibrated probabilistic predictions for the specific locations or areas (not grid cells) which are of interest to stakeholder decision making \citep{beven2015hyperresolution}. However, the issue is not simply one of scale, and increasing the resolution of imperfect models does not solve the problem of what \citet{beven2015hyperresolution} term `hyperresolution ignorance' in that uncertainty about parameters will still exist, and a model outputting at finer scales will not necessarily be providing more information. This is a topical issue for weather forecasting too, as numerical weather prediction models continue to increase in resolution.

Technological challenges such as limited computational power have slowed the adoption of the `models of everywhere' concept, but \citet{blair2019data} propose that data science, including cloud computing infrastructure, may provide the means to make the `models of everywhere' concept a reality by using data mining techniques to combine information from remote sensing and in-situ earth monitoring systems in data-driven models. This includes live-updating IoT sensors \citep{atzori2010internet, nundloll2019design} such as the unofficial weather stations which provide data to the Met Office's Weather Observation Website \citep{kirk2021weather}. These provide a greater number of observations from more numerous unique sites than traditional monitoring systems (e.g. traditional weather stations), and the combination of increasingly dense observations (by IoT sensors) and machine learning may allow data-driven models to supersede alternative modelling approaches, such as `generic' physics based models. However, IoT sensors have greater potential for data quality issues over traditional monitoring systems owing to their cheaper costs, less stringent maintenance, and being more numerous. Therefore, in order to maximise the benefits of IoT sensor data for environmental modelling, issues of data quality have to be addressed as part of the solution. 

We propose that the approach we present here, which uses Bayesian deep learning to combine information from remote sensing and in-situ earth monitoring in order to provide specific and well-calibrated predictions for any point within the extent of observed space and time, does satisfy the ideals behind the `models of everywhere' concept. As such, it can be viewed as an example of the kind of large scale data-driven environmental modelling that is likely to become more feasible as computing power continues to increase - putting `models of everywhere' at our fingertips.

\begin{figure}[htb]
    \centering
  \begin{subfigure}[t]{0.02\textwidth}
    \textbf{A}
  \end{subfigure}
  \begin{subfigure}[t]{0.47\textwidth}
    \includegraphics[width=\linewidth, valign=t]{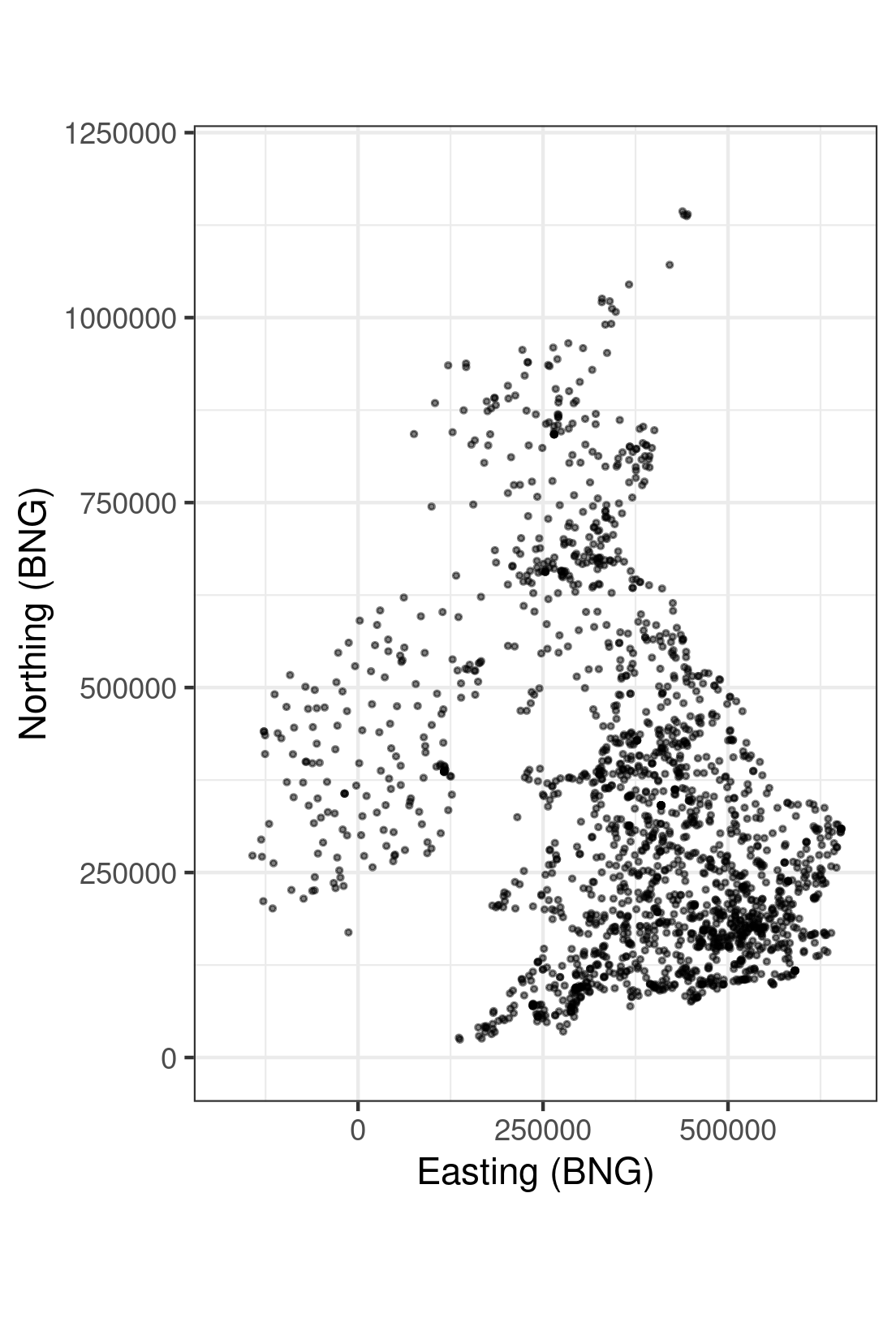}
  \end{subfigure}\hfill
  \begin{subfigure}[t]{0.02\textwidth}
    \textbf{B}
  \end{subfigure}
  \begin{subfigure}[t]{0.47\textwidth}
    \includegraphics[width=\linewidth, valign=t]{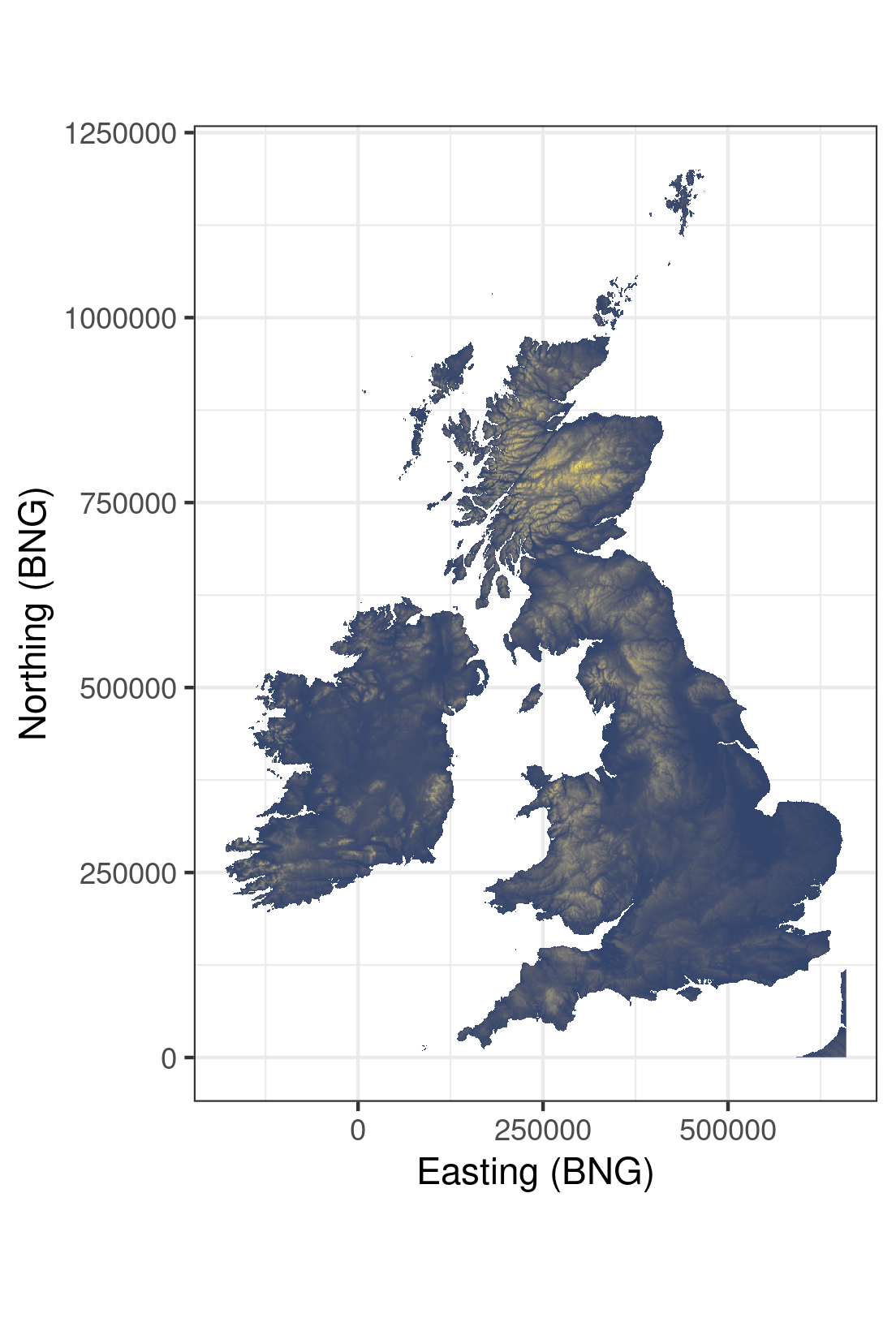}
  \end{subfigure}
    \caption{\textbf{A}, The locations of the 1893 private weather stations across the British Isles that provide crowd-sourced data to the the Weather Observation Website. \textbf{B}, SRTM elevation data for the British Isles which our model uses as auxiliary information.} 
    \label{fig:britishisles}
\end{figure}

\begin{figure*}[htb]
    \centering
    \includegraphics[width=\textwidth]{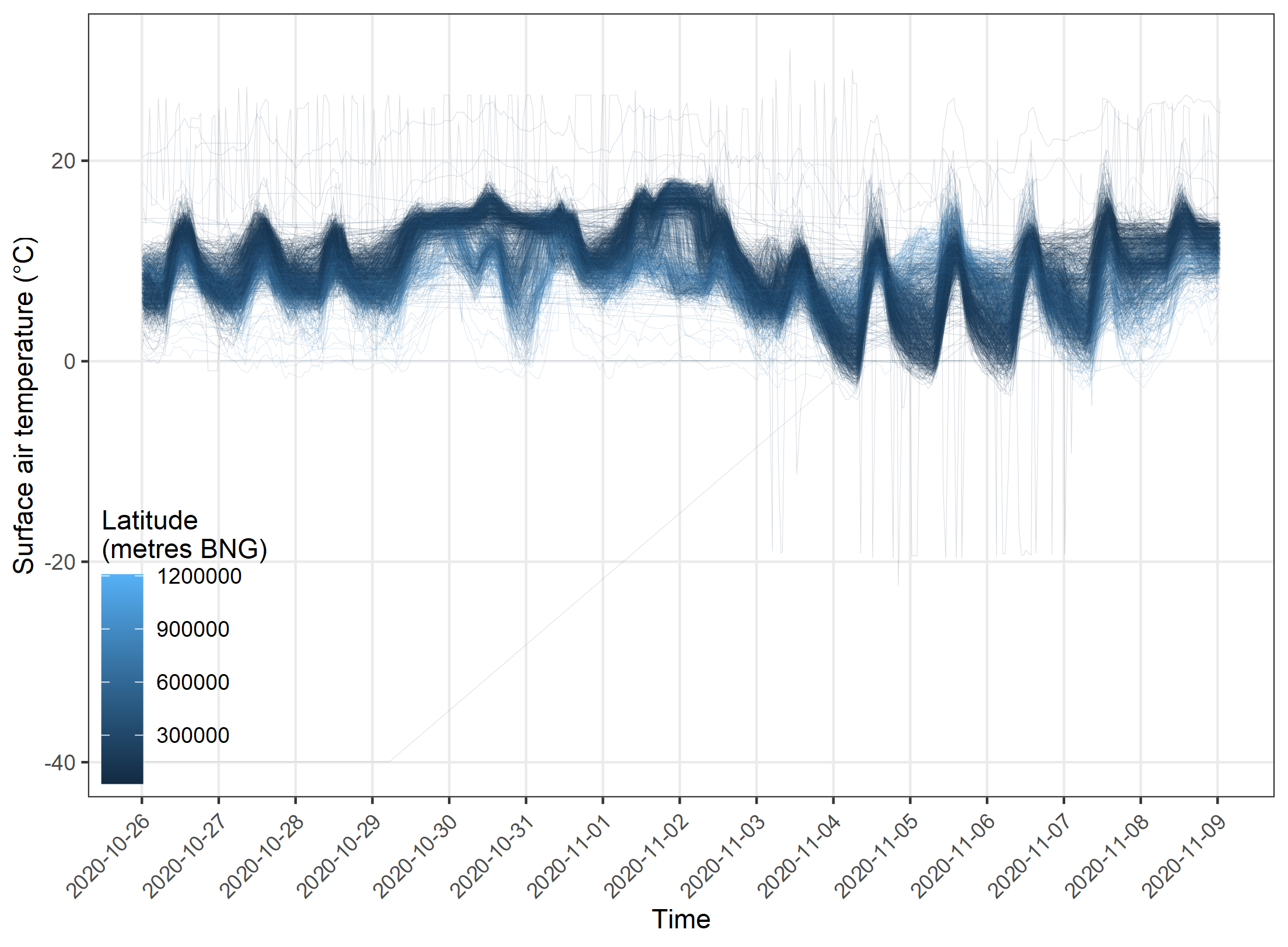}
    \caption{Time-series visualisation of our Met Office Weather Observation Website temperature data for a seven day period in November 2020. Each line represents one weather station, which are coloured such that higher latitudes are a lighter shade. Note how the data is not clean, but contains spurious observations on either side of the central distribution}
    \label{fig:timeseries}
\end{figure*}

\subsection*{Outlier detection}

There are many possible approaches to outlier detection, ranging from fully-manual data checking, to manually designed rule-based filters, to statistical and machine learning based systems, which may include both supervised and unsupervised learning \citep[with supervised learning having the downside that it requires the creation of manually labelled training datasets in advance; e.g.][]{nesa2018outlier}. For a full review of outlier detection techniques we refer the reader to \citet{wang2019progress}, who provide a general review of developments in outlier detection since the year 2000. In addition, \citet{ayadi2017outlier} provide a review of techniques specifically for wireless sensor networks, including a comparison of the respective pros and cons of statistical, nearest-neighbour, artificial intelligence, clustering and classification based approaches (although these categories have some overlap). \citet{napoly2018development} propose a combination of rule-based and z-score thresholds for outlier detection in crowdsourced air temperature data. This approach has been adopted by other authors \citep[e.g.][]{venter2020hyperlocal, zumwald2021mapping} but this is not the approach we take.

The approach we adopt for this study is a regression approach using a deep neural network mixture model --- or mixture density network \citep{bishop1994mixture} --- through which we represent the conditional distribution of reported temperature values as a mixture of a Gaussian and a Uniform distribution, with parameters learned by our deep neural network. We explain the full details of the approach in subsequent sections, but in brief terms, our approach incorporates outlier detection into the spatio-temporal modelling process itself, by having the neural network learn the probability that an observation is an outlier (whose values are best explained as having been generated by the Uniform distribution) as an unsupervised sub-task to the overall supervised spatio-temporal modelling task. The benefit of this holistic approach is that it allows the user to incorporate knowledge about data issues into the data model itself through the use of suitable probability distributions, and makes for more seamless model checking when compared to a two-stage procedure of separate outlier-detection followed by data modelling. 

\section*{Method}
\subsection*{Dataset}

We demonstrate our approach using surface air temperature data from the Met Office's Weather Observation Website archives \citep{kirk2021weather}. These data contain observations from 1893 unique IoT type weather stations (\autoref{fig:britishisles}), from which we have taken a continuous 14 day window from 2020/01/26 to 2020/11/09 to use as our dataset in this study. The data provide our target variable, surface air temperature in degrees Celsius, as well as spatio-temporal location information in the form of British National Grid (BNG) Easting and Northing, and a timestamp. Collectively these Weather Observation Website weather stations record 8000 observations per hour on average, which equates to about four observations per site per hour, although this varies by site. Each sensor records observations at different intervals, rather than synchronously at set times, so that collectively the observations provide good coverage across continuous time (\autoref{fig:timeseries}).

In addition to using the Weather Observation Website data, we also make use of gridded UK elevation data as covariate or auxiliary information in order to help inform the spatio-temporal interpolation. The data used comes from NASA's Surface Radar Topography Mission \citep[SRTM;][]{farr2007shuttle} and is accessed via the Raster package in the R programming language. The elevation data is rasterised with a grid size of 528 by 927 metres (longer latitudinally than longitudinally), resulting in 0.66 million grid cell elevation dataset covering the UK and Ireland.  

For input into our model, we extract terrain elevation images centred on each observation (in the case of training) or location to be predicted. The images extracted have a resolution of 32x32 grid cells with a grid cell size of 500m (we use bilinear interpolation so that the image resolution is not locked to the overall digital elevation model resolution). These images provide auxiliary information, from which the convolutional layers of our deep neural network learn to extract useful contextual covariates \citep[e.g. as explained in][]{kirkwood2020deep} for the task of spatio-temporal interpolation of surface air temperature data. Illustrative examples could include slopes facing the sun that warm faster, or valleys that channel cool air from cold mountainous areas. There are likely to be many such complex interactions between the landscape and surface air temperatures, and by providing elevation data as images to our deep neural network we allow them to be learned from data. Further details of the preparation of our dataset for model training, evaluation, and testing are provided in the section `Practical setup'.   

\subsection*{Mixture model concept}

We design our model to address three considerations: 1) The capacity to represent our target phenomenon (a spatio-temporally varying temperature distribution in this case), under the assumption that outliers can be objectively identified and excluded. 2) The capacity to successfully identify outliers. 3) A means by which to achieve both 1 and 2 simultaneously within a single probabilistic data model.

At the heart of our model is a two-part mixture probability distribution whose individual component distributions - $p_{signal}$ and $p_{outlier}$ - represent the two classes of observation that we judge to exist within our dataset, as evidenced by exploratory visualisation of the data (\autoref{fig:timeseries}). These are 1) The `true' signal distribution of our target phenomenon, which we assume here is a Gaussian distribution as is common for temperature measurements, and 2) the outlier distribution, in this case we choose a Uniform distribution `catch-all' that can account for the generation of spurious observations by biased or faulty weather stations. It is worth noting that the selection of these distributions is a modelling choice, and that different target variables are likely to warrant the use of different distributions in the model output, from which the likelihood is derived (the probability of the data given the model).

We then introduce parameter $\theta$ -- the probability that an individual data point comes from the ``true'' Gaussian distribution of temperature. Equivalently, $1-\theta$ is the probability that a data point is spurious and therefore comes from the uniform distribution. More formally, let $y_{s,t}$ denote the temperature at location $s$ and time point $t$. The probability distribution of $y_{s,t}$ is defined as:
\begin{eqnarray}
p(y_{s,t}) &=& \theta_{s,t} p_{signal}(y_{s,t}) + (1-\theta_{s,t})p_{outlier}(y_{s,t}), \label{mixture1} \\
p_{signal}(y_{s,t}) &=& \mbox{Normal}(\mu_{s,t},\sigma^2_{s,t}) \mbox{ with density } \frac{1}{\sqrt{2\pi}\sigma_{s,t}}\exp\left\{\frac{1}{2\sigma_{s,t}^2}(y_{s,t}-\mu_{s,t})^2\right\}  \label{mixture2} \\
p_{outlier}(y_{s,t}) &=& \mbox{Uniform}(\mu_{s,t} - 50, \mu_{s,t} + 50) \mbox{ with density } \frac{1}{100}. \label{mixture3}
\end{eqnarray}
The ``true'' temperature distribution is therefore assumed Normal with mean $\mu_{s,t}$ and variance $\sigma^2_{s,t}$, while the spurious observations are centered at the ``true'' mean $\mu_{s,t}$ but are allowed to vary uniformly around this mean. This range of 100$^\circ$C was chosen from exploratory data analysis and was deemed sufficient to capture the outliers in the data.

A perhaps more intuitive way of interpreting this model, is to introduce a latent binary variable, $Z_{s,t}$ where $\mbox{Pr}(Z_{s,t}=1)=\theta_{s,t}$ and $\mbox{Pr}(Z_{s,t}=0)=1-\theta_{s,t}$. The probability model for temperature $y_{s,t}$ conditional on $Z_{s,t}$ is then:
\begin{eqnarray}
y_{s,t}|Z_{s,t}=1 &\sim& \mbox{Normal}(\mu_{s,t},\sigma^2_{s,t}) \\
y_{s,t}|Z_{s,t}=0 &\sim& \mbox{Uniform}(\mu_{s,t} - 50, \mu_{s,t} + 50) \\
Z_{s,t} &\sim& \mbox{Bernoulli}(\theta_{s,t}).
\end{eqnarray}
We can think of $Z_{s,t}$ as the result of a `coin toss' where at any given location $s$ and time point $t$, we can get a spurious observation with probability $1-\theta_{s,t}$. Note that $\theta_{s,t}$ varies with space and time, in order to flexibly capture the flawed data points (as opposed to assuming a constant $\theta$).

Note further that the fact that $\mu_{s,t}$ appears in the model for the spurious data points, i.e. the Uniform distribution, allows some information from such data points to be utilised. This is based on a belief that on average, the flawed data are centred on the true mean, such that negative and positive biases cancel each other out \citep[though in practice this may well be optimistic;][]{bell2015good}. Note that any flawed data point which is much further from the mean $\mu_{s,t}$ than the $\mbox{Normal}(\mu_{s,t},\sigma^2_{s,t})$ distribution implies, will be ``absorbed'' by the Uniform distribution so that the Normal part of the model can be interpreted as the model for the true temperature process. As such, predictions from the $\mbox{Normal}(\mu_{s,t},\sigma^2_{s,t})$ part after the model is implemented, can be seen as ``corrected''.

\begin{figure*}[!htb]
    \centering
    \includegraphics[width=\textwidth]{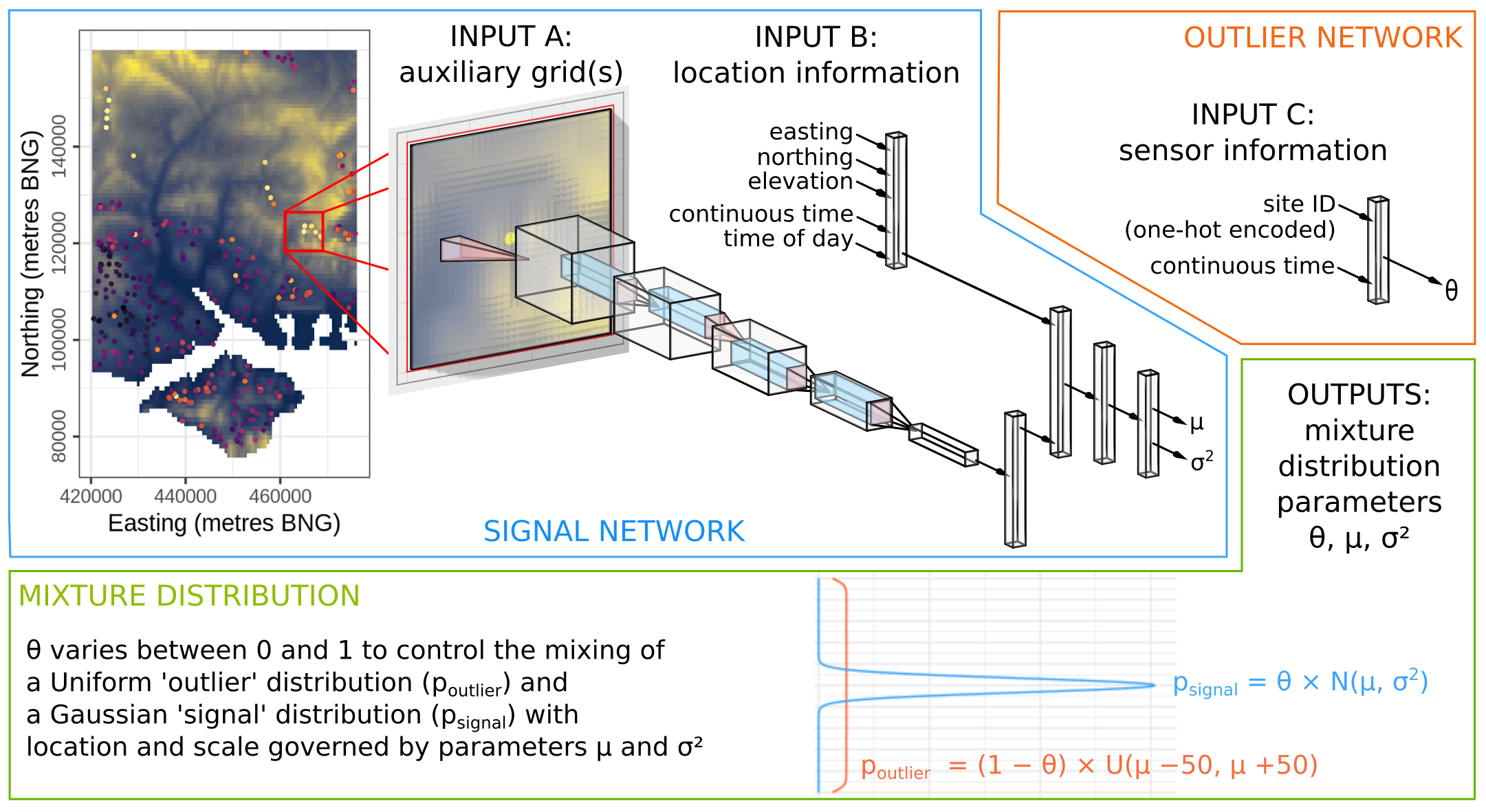}
    \caption{The architecture of our deep neural network, which combines a signal modelling network and an outlier detection network. The signal network learns the parameters of our Gaussian output distribution as a function of its inputs for each observation. Meanwhile the outlier network learns the probability of an observation being an outlier as a function of site ID and time (based on the likelihood of the observation having been generated by the Uniform distribution rather than the Gaussian). Architecture extends from \citet{kirkwood2020bayesian}.}
    \label{fig:architecture}
\end{figure*}

\subsection*{Network architecture}

The parameters of our mixture distribution are $\mu_{s,t}$, $\sigma^2_{s,t}$ and $\theta_{s,t}$. We therefore require that our model has the capacity to learn to optimise these parameters in relation to space and time so that predictions from \eqref{mixture2} are a reasonable representation of the real data generating processes at location $s$ and time $t$ (as we assess through model checking against held-out test data).

To achieve this, our neural network architecture consists of two halves, which we term the signal network and the outlier network. The signal network is tasked with learning the parameters, $\mu_{s,t}$ and $\sigma_{s,t}^{2}$, of our `true' Gaussian distribution, which are conditioned on the space and time variables that we provide as inputs to the model (the details of which we explain in subsequent sections). The outlier network meanwhile is simply tasked with learning $\theta_{s,t}$ or in other words the probability that an observation is an outlier, which is conditioned on site ID (which we provide one-hot encoded) and time. One-hot encoding means representing our n site IDs as n separate predictor variables, to which we assign the value 1 only if an observation corresponds to that site, otherwise a value of 0 is assigned. The one-hot encoding approach allows us to input categorical variables into the neural network in a sensible way. We provide site IDs (rather than more general spatial variables such as easting and northing) to the outlier network because it has no need to learn generalisable patterns, its sole purpose is to identify outliers probabilistically during the training phase, and this ability is improved by making the task as simple as possible. Overfitting is not a concern since the outlier network serves no purpose in the spatio-temporal interpolation beyond the training stage.

From the perspective of deep neural networks as ``black boxes'', we can view our signal and outlier networks simply as function approximators that learn to provide optimal values of their respective output parameters, such that

\begin{eqnarray}
(\mu_{s,t}) &=& g(f_{signal network}(x_{space,time})),\label{signalmu}\\
(\sigma^{2}_{s,t}) &=& h(f_{signal network}(x_{space,time})),\label{signalsigma}
\end{eqnarray}

and

\begin{equation}\label{outlier}
\mbox{logit}(\theta_{s,t}) = f_{outlier network}(x_{site,time}),
\end{equation}

however we have designed the architecture of the two branches --- signal network and outlier network ---  in line with their specific goals. The architectures of each branch, and the specific space and time variables that they take as inputs, are explained in the following paragraphs, accompanied by \autoref{fig:architecture} as a visual aid.

Our signal network architecture (\autoref{fig:architecture}) is designed for terrain-aware interpolation, which it achieves through the combination of a convolutional branch to derive relevant terrain features from gridded auxiliary information (e.g. terrain elevation, satellite imagery), and a fully-connected branch for interpolation in space and time. The combined effect is to achieve spatio-temporal interpolation in a hybrid space that includes local terrain context so that, for example, the differences between valleys and hill-tops (and anything relevant about their orientations) can be recognised. Unlike more traditional geostatistical approaches, which might offer the model pre-defined derivatives from terrain analysis as input features, our deep learning approach allows these derivatives to be learned optimally for the task at hand via trainable convolutional filtering of raw terrain elevation grids \citep{behrens2018multi, padarian2019using, wadoux2019using, kirkwood2020deep, kirkwood2020bayesian}.

For its location input (input B in \autoref{fig:architecture}) our signal network receives easting, northing, and elevation as spatial location information (all in metres), and continuous time and time of day as temporal location information (in minutes). To provide a cyclic representation of time of day to the network (to aid learning of the diurnal cycle), we transform our minute-of-the-day variable into position on a circle defined by the two dimensions $\sin(2\pi t/T)$ and $\cos(2\pi t/T)$ where $t$ is the specific minute of the day and $T = 1440$: the total number of minutes in a day. It is important that our signal network is able to generalise well to unobserved locations, and so over-fit must be avoided. In aid of this, and in line with the Bayesian interpretation of our model --- which we discuss in the next section --- we run our signal network with a dropout rate of 0.5 on all hidden layers (or spatial dropout in the case of convolutional layers). 

In contrast, generalisation is not a concern for our outlier network (\autoref{fig:architecture}), whose sole task is to model the probability that each training observation is an outlier. To make this task as simple as possible, we provide the outlier network directly with one-hot encoded site IDs, as well as continuous time, such that the outlier network provides outlier probabilities as a linear function of site ID plus a (site-tailored) non-linear function of continuous time (eq.~\ref{outlier_explan}) facilitated by passing continuous time through a single hidden layer.

\begin{equation}\label{outlier_explan}
\mbox{logit}(\theta_{s,t}) = f_{outlier network}(x_{site,time}) \approx \beta_{SITE_s} + f_{SITE_s}(t)
\end{equation}

For full layer-by-layer details of our neural network architecture, we encourage readers to view our code for this study at https://github.com/charliekirkwood/deepoutliers.

\subsection*{Bayesian inference}

With the parameters and architecture of our model established, we would like to use the Bayesian framework to learn a posterior distribution for all trainable parameters given the data, $D$, on which we will train the model. The parameters that control the probability distribution of temperature are $\theta_{s,t}$, $\mu_{s,t}$ and $\sigma^2_{s,t}$ but these are of course themselves functions of the weights within the entire neural network, which we collectively refer to as $w$. By Bayes' rule we can obtain this posterior distribution over the weights given the data as

\begin{equation}\label{bayes}
 p(w|D) = \frac{p(D|w)p(w)}{p(D)}
\end{equation}

So that $p(w|D)$ is proportional to the likelihood of the data given the weights, $p(D|w)$, multiplied by our prior distribution over the weights, $p(w)$. Assuming independence of temperature values given $\theta_{s,t}$, $\mu_{s,t}$ and $\sigma^2_{s,t}$, the likelihood of our mixture model is:
\begin{equation}
p(D|w) = \prod_s \prod_t p(y_{s,t})
\end{equation}
where $p(y_{s,t})$ is given in equation \eqref{mixture1}.

Here, we adopt a prior distribution for $w$ by utilising Monte Carlo Dropout as suggested by \cite{gal2016dropout}. The prior is defined by assuming that a particular ``fixed'' weight $\beta_j$ in the network can be randomly ``dropped out'', by introducing a set of Bernoulli random variables $B_j\sim\mbox{Bern}(\pi)$. An individual weight $w_j$ is then defined as
\begin{equation} \label{prior}
w_j = B_j\beta_j 
\end{equation}
so that $w_j=\beta$ with probability $\pi$ and $w_j=0$ with probability $1-\pi$. The fixed weights $\beta$ are learned by stochastic gradient descent during training, whereas the dropout rate $\pi$ is considered a hyper-parameter of the network and is fixed a-priori. Equation \eqref{prior} means that the weights $w_j$ are probabilistic in nature so that stochastic forward passes can be used in a Monte Carlo setting to provide an approximate posterior distribution $p(w|D)$ for $w$.

The particular setup assumes that $\pi$ is fixed a-priori, preferably by tuning it. It is however possible that this is automatically estimated using `Concrete Dropout' \citep{gal2017concrete}, or by exploring the number of other approaches to Bayesian inference in neural networks that have been proposed \citep[e.g.][]{mackay1995probable,graves2011practical,neal2012bayesian,heek2019bayesian}. At present, Bayesian inference in deep neural networks, with their extreme dimensionality (a modest 696 114 trainable parameters in our case), remains a challenge and an ongoing topic of research.

After obtaining the posterior distribution $p(w|D)$ (the practicalities of which we discuss in the next section), we are in a position to compute the posterior predictive distribution for any point in space and time. To obtain robust predictions of the phenomenon of interest, we can set $\theta_{s,t}=1$ (i.e. exclude the uniform distribution component) and thus generate predictions exclusively from the `true' Gaussian distribution \eqref{mixture2}. Specifically, we can obtain samples from the posterior predictive distribution of any $y_{s,t}$ (both observed and not):
\begin{equation}\label{predictive}
 p(y_{s,t}|D) = \int_{w}^{}p_{signal}(y_{s,t}|w)p(w|D)dw,
\end{equation}

\subsection*{Practical setup}

We use Weather Observation Website surface air temperature observations from a fourteen day period from 26/10/2020 to 09/11/2020 for this study. This period was selected for containing interesting weather patterns (as evident even in the simple time series of observations; \autoref{fig:timeseries}), including storm Aiden which passed over on the UK on the 31st of October 2020. We randomly subsampled the observations from this period to a single observation per site per hour (where available), which provides 417141 observations in total. We then split this dataset by site ID into 10 folds of approximately equal unique number of unique sites (about 145 unique sites per fold). We split our folds in this site-wise manner in order to assess the fit of the model at sites unseen during training, and therefore to assess the ability of the model to interpolate to new spatial locations throughout the period of observed time. 

We assigned data folds one to eight to be used for training, with fold nine providing an evaluation set for hyper-parameter tuning, and fold ten providing a held out test set for assessing the performance of the final trained model at locations unseen by the model. Running on a single GPU workstation (with Nvidia GTX 3070) our neural network trains at one epoch every 3 seconds, so that training for 600 epochs takes about 30 minutes.

\section*{Results and discussion}

\begin{figure*}[!htb]
    \centering
    \includegraphics[width=\textwidth]{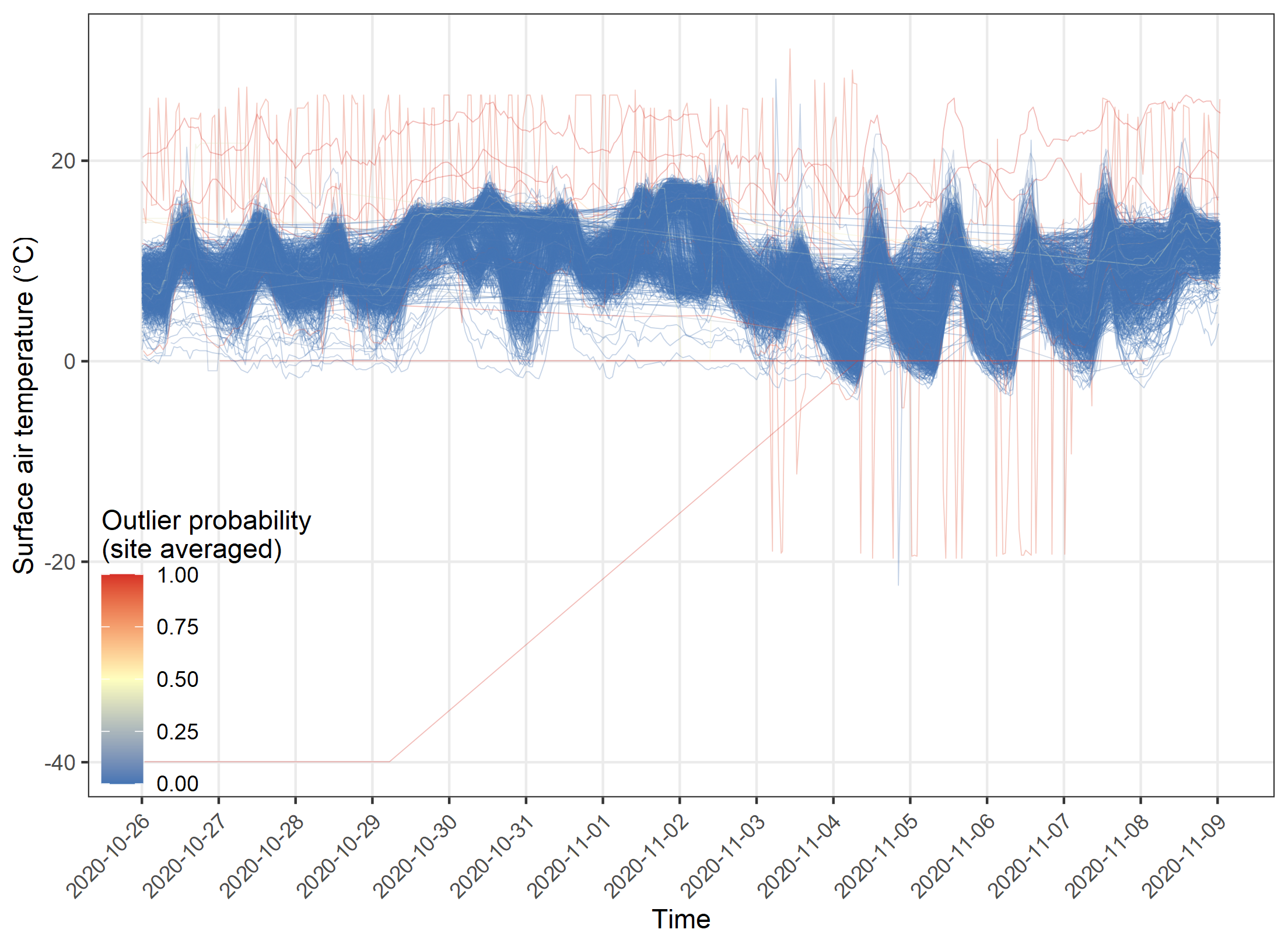}
    \caption{Time-series visualisation of our training dataset, coloured by posterior probability that observations would more likely be generated by the uniform outlier distribution rather than the Gaussian signal distribution (for the purposes of the figure, probabilities are averaged by site in order to obtain a fixed line colour per site). We proceed to make predictions using only the Gaussian output of our deep mixture model, such that all predictions represent the true, clean temperature field without spurious measurements.}
    \label{fig:outliers}
\end{figure*}

Our approach has not required the manual labelling of outliers in the training data, but we can see from the output of the model --- specifically the parameter $\theta$, which controls the mixing of the Gaussian and Uniform distributions --- that observations that visually appear to be outliers have been assigned a high probability of being outliers generated by the Uniform distribution (see for example \autoref{fig:outliers}, in which sites are coloured by the average predicted outlier probability of their observations). On the basis of this qualitative assessment, we have confidence that predictions generated by our neural network's Gaussian output distribution are a clean (outlier free) representation of the true surface air temperature - we also find this to be evident in the clean look of maps generated by the model (using only the Gaussian distribution for prediction), which do not contain the localised bright or dark spots that could be expected if the model had incorrectly fitted to outlier observations. All subsequent reporting of results, and their discussion, is made on the basis of using only the Gaussian distribution for prediction, so that all predictions are `outlier-filtered'.

\begin{figure}[!htb]
    \centering
    \begin{subfigure}[t]{0.02\textwidth}
    \textbf{A}
  \end{subfigure}
  \begin{subfigure}[t]{0.47\textwidth}
    \includegraphics[width=\linewidth, valign=t]{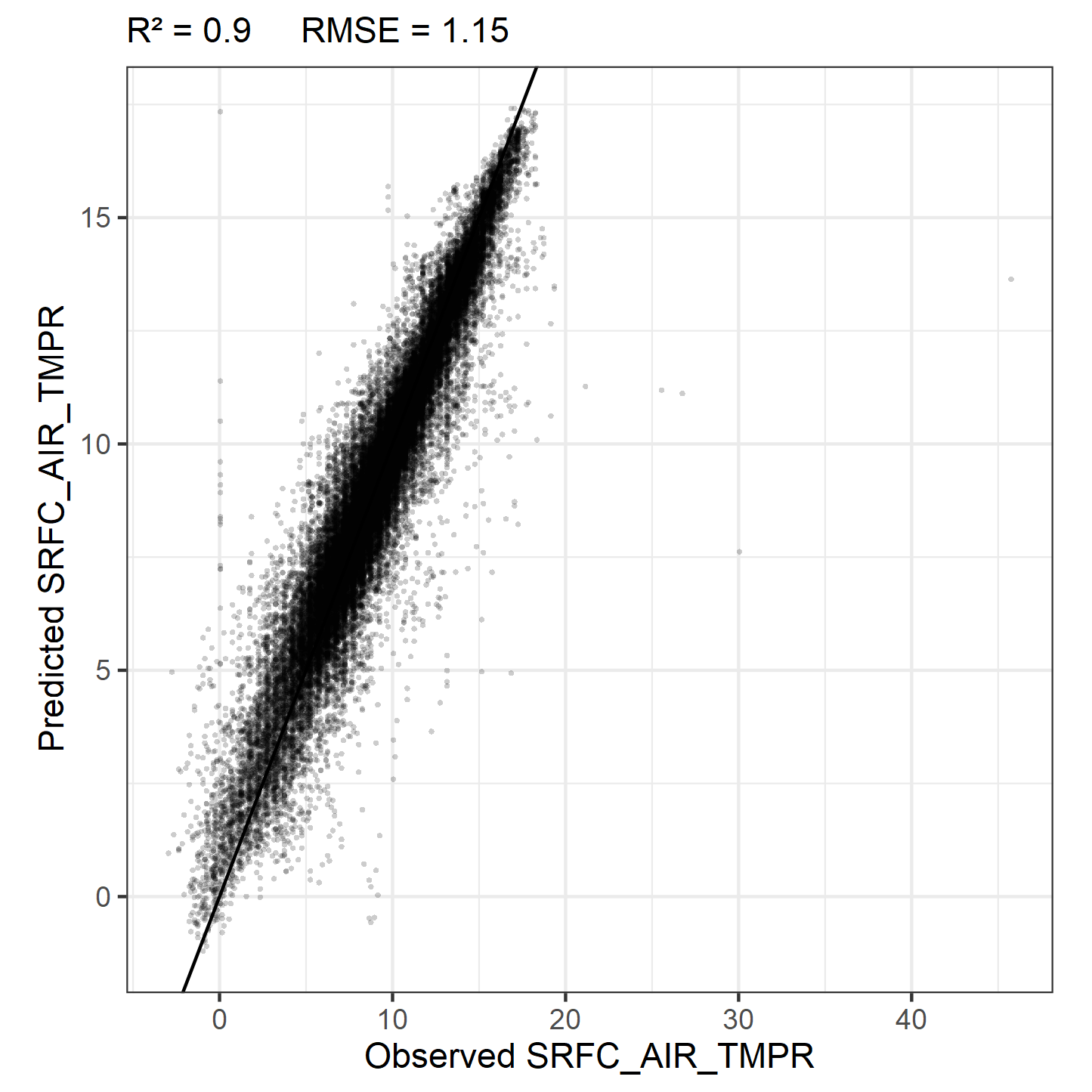}
  \end{subfigure}\hfill
  \begin{subfigure}[t]{0.02\textwidth}
    \textbf{B}
  \end{subfigure}
  \begin{subfigure}[t]{0.47\textwidth}
    \includegraphics[width=\linewidth, valign=t]{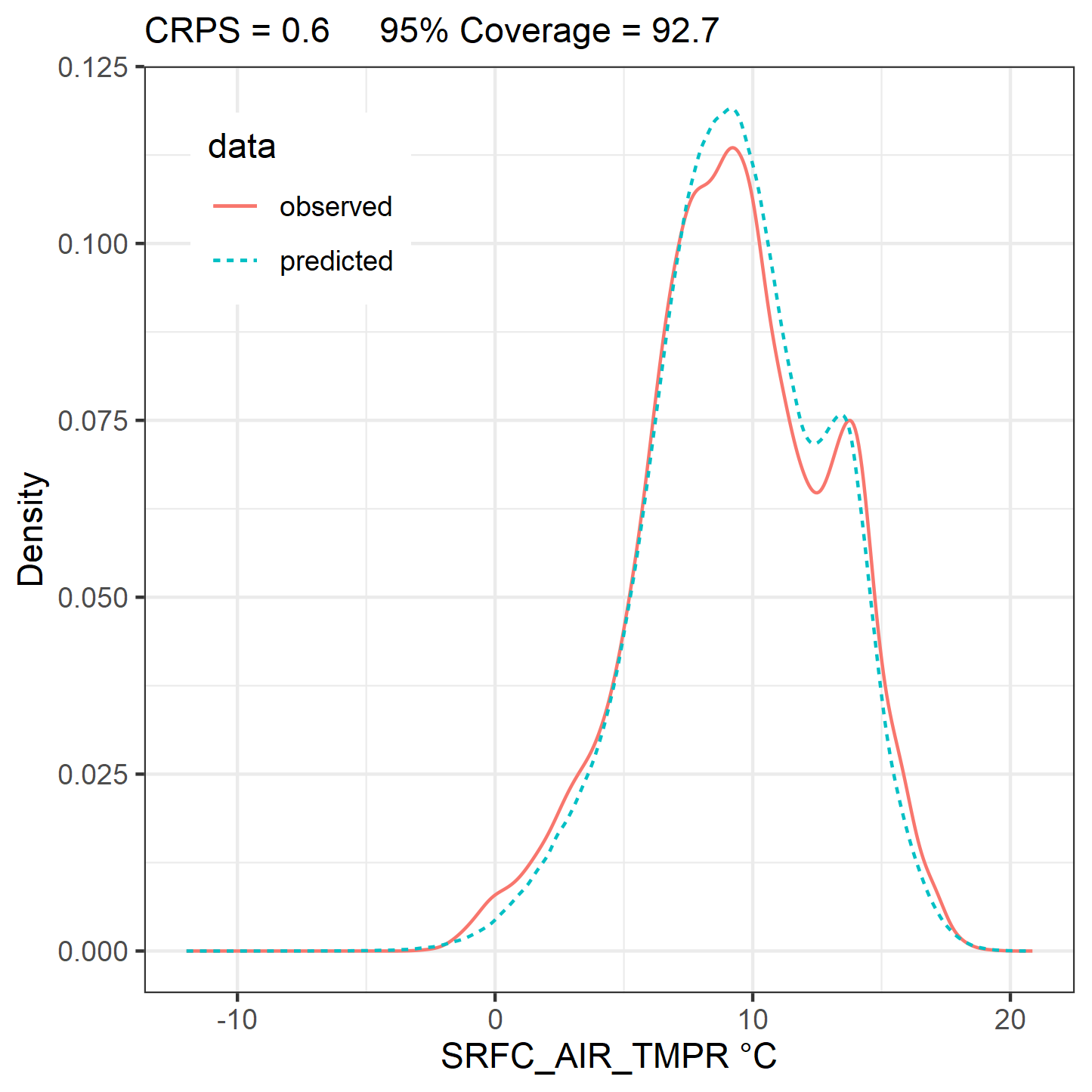}
  \end{subfigure}
      \centering
    \begin{subfigure}[t]{0.02\textwidth}
    \textbf{C}
  \end{subfigure}
  \begin{subfigure}[t]{0.47\textwidth}
    \includegraphics[width=\linewidth, valign=t]{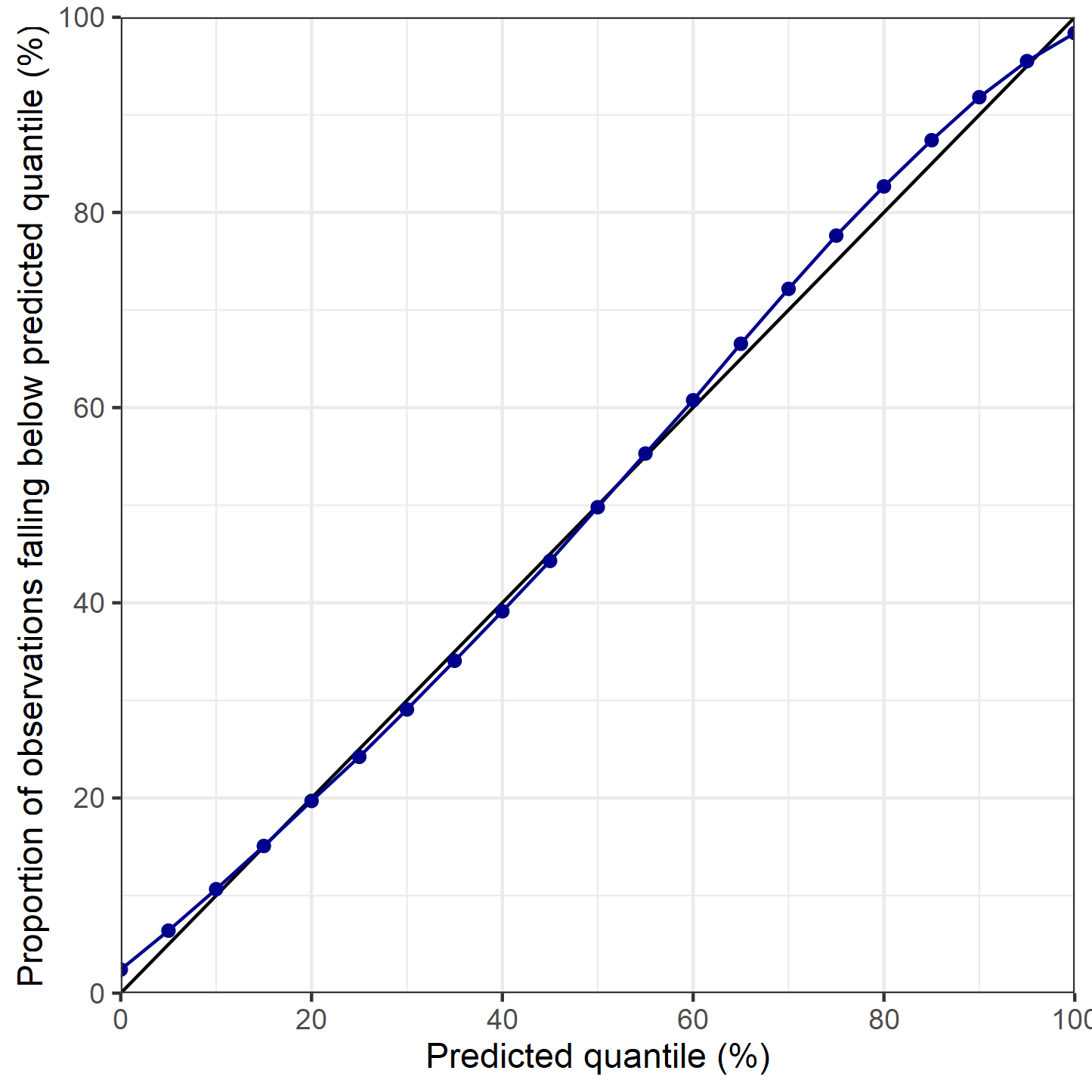}
  \end{subfigure}\hfill
  \begin{subfigure}[t]{0.02\textwidth}
    \textbf{D}
  \end{subfigure}
  \begin{subfigure}[t]{0.47\textwidth}
    \includegraphics[width=\linewidth, valign=t]{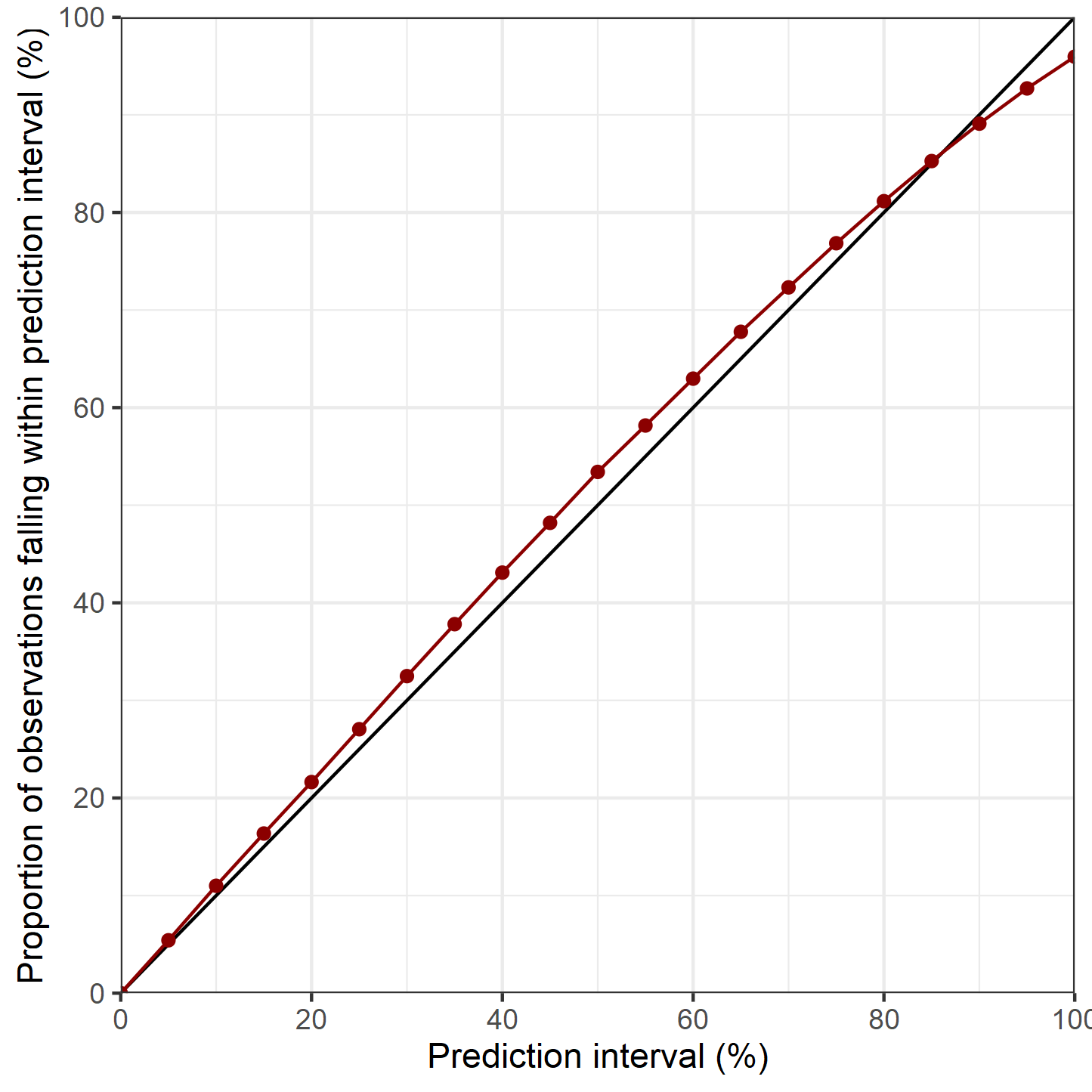}
  \end{subfigure}
      \caption{\textbf{A}, Deterministic comparison of observed and predicted values of surface air temperature, taking the mean of the predictive distribution as the prediction. Note some outlier observations are present in the test set. \textbf{B}, Probabilistic comparison of observed and predicted distributions of surface air temperature, taking 50 samples from the predictive distribution for each observation. \textbf{C}, Q-Q plot and \textbf{D}, prediction interval coverage plot to check the calibration of our model's predictive distribution against test observations.   
      All use data from the held out test set (n = 41836), taken from sites kept unseen until after hyper-parameter tuning and model training.}
    \label{fig:metrics}
\end{figure}

In terms of the quantitative performance of the model as assessed on held out test data (from sites unseen by the model during training), we find that our deep learning approach to spatio-temporal interpolation provides a good degree of predictive skill in both a deterministic and probabilistic sense (\autoref{fig:metrics}). In a deterministic sense (\autoref{fig:metrics}A), the mean of the predictive distribution provides an R$^2$ of 0.90 and a root mean square error (RMSE) of 1.15 degrees Celcius. Probabilistically, our model achieves a continuous rank probability score of 0.6 (\autoref{fig:metrics}B), and the predictive distribution has good calibration, with held out test observations falling within the 95\% prediction interval 92.7\% of the time. We can see from the quantile-quantile plot (\autoref{fig:metrics}C) and prediction-interval coverage plot (\autoref{fig:metrics}D) that the probabilistic calibration of the predictive distribution performs well across the range of predicted quantiles, although we do see a slight under-dispersion in the tails (i.e. beyond a 90\% prediction interval). This may be attributable to limitations of our Monte Carlo dropout approach to approximate Bayesian inference, in that our posterior distribution is fundamentally centred about a single optimum, rather than composed of diverse samples from separate local optima as in full Bayesian inference via MCMC sampling methods \citep[or other proposed approximations such as the `deep ensembles' approach;][]{lakshminarayanan2016simple} which may reduce the diversity and coverage of the posterior. However, as we assess here on held-out test data, this under-dispersion, if present, appears to be minimal and not overly concerning, especially given that some of our test observations are outliers themselves, which means that perfect calibration (using predictions from our Gaussian distribution alone) cannot be expected.

Overall the performance metrics indicate that our deep mixture density network approach to outlier-filtered spatio-temporal interpolation is doing a good job of providing accurate and trustworthy predictions of historic surface air temperatures for locations in space which have not been observed. It provides a statistical hindcast which is likely to be both computationally cheaper and better calibrated than numerical hindcast alternatives. When run over a long duration, our approach should also provide high quality probabilistic climatology estimates at any unobserved location, which may be useful for planning purposes.

\begin{figure*}[!htb]
    \centering
    \includegraphics[width=\textwidth]{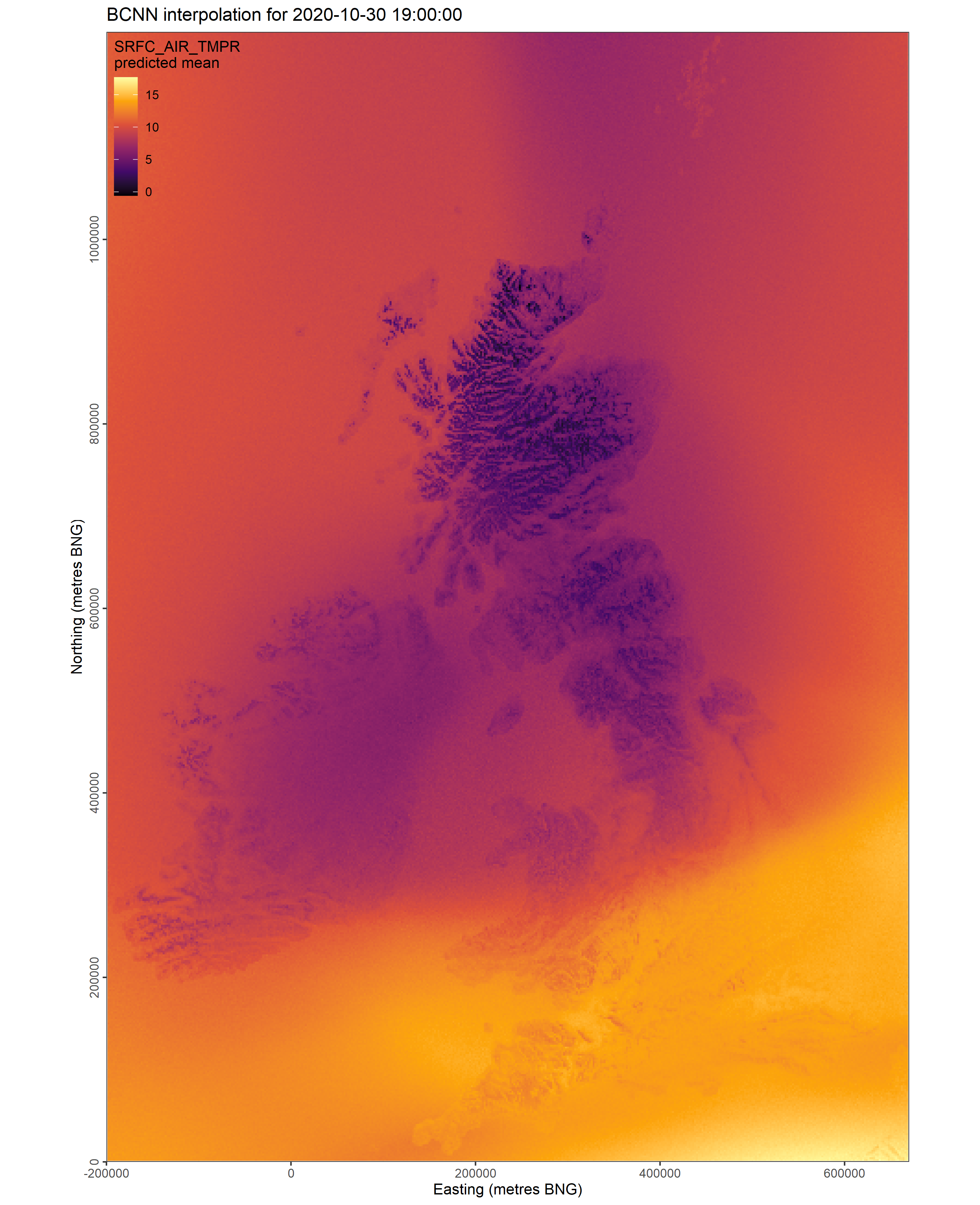}
    \caption{Mean surface air temperature map for a single snapshot in time, as predicted by our deep neural network. The use of convolutional layers in our neural network architecture allows our predictions to be informed by patterns relating air temperature to terrain features, and in doing so produce detailed spatio-temporal fields.}
    \label{fig:meanmap}
\end{figure*}

\begin{figure*}[!htb]
    \centering
    \includegraphics[width=\textwidth]{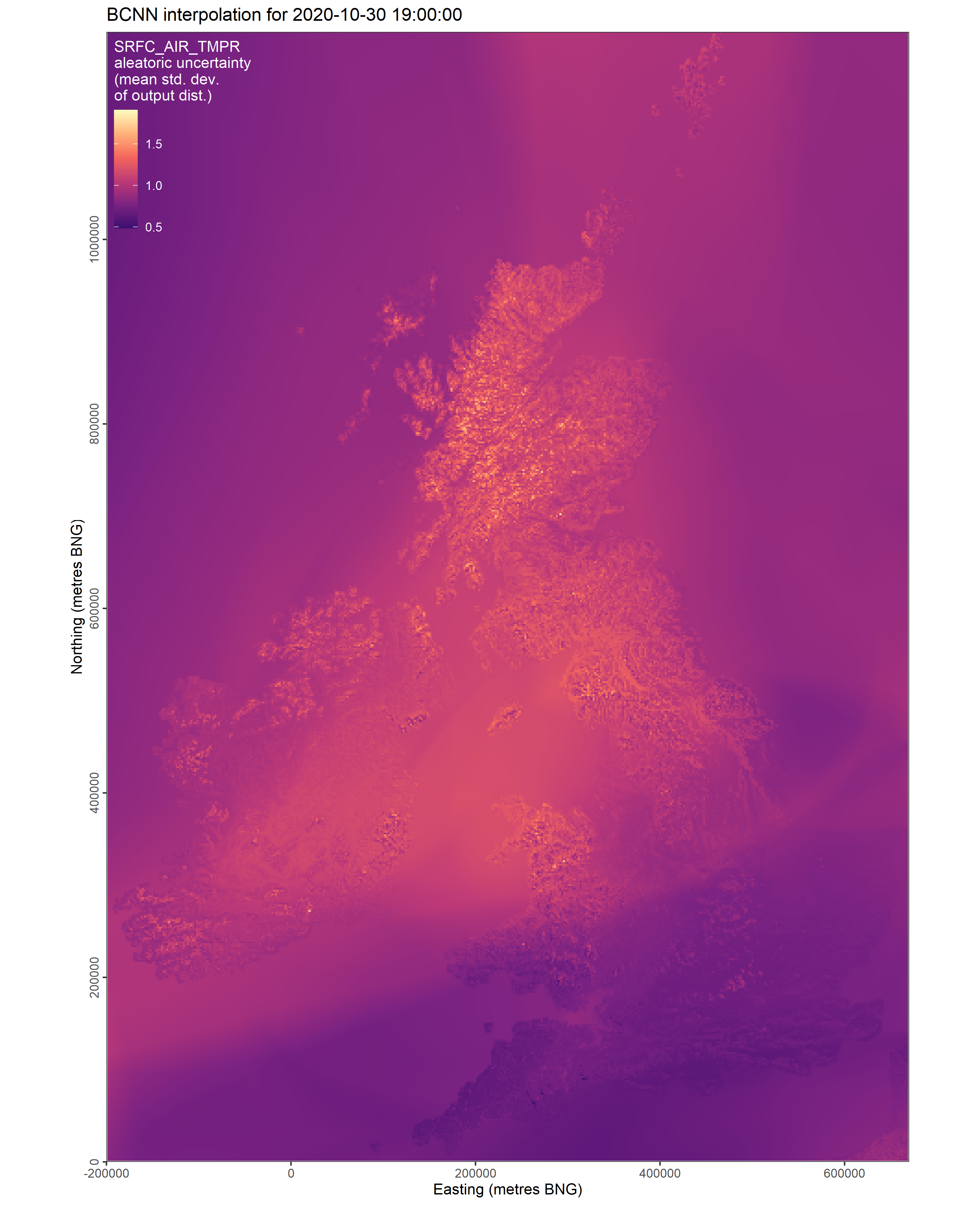}
    \caption{Aleatoric uncertainty (mean standard deviation of the Gaussian output distribution, °C) at the same snapshot in time.}
    \label{fig:aleatoricmap}
\end{figure*}

\begin{figure*}[!htb]
    \centering
    \includegraphics[width=\textwidth]{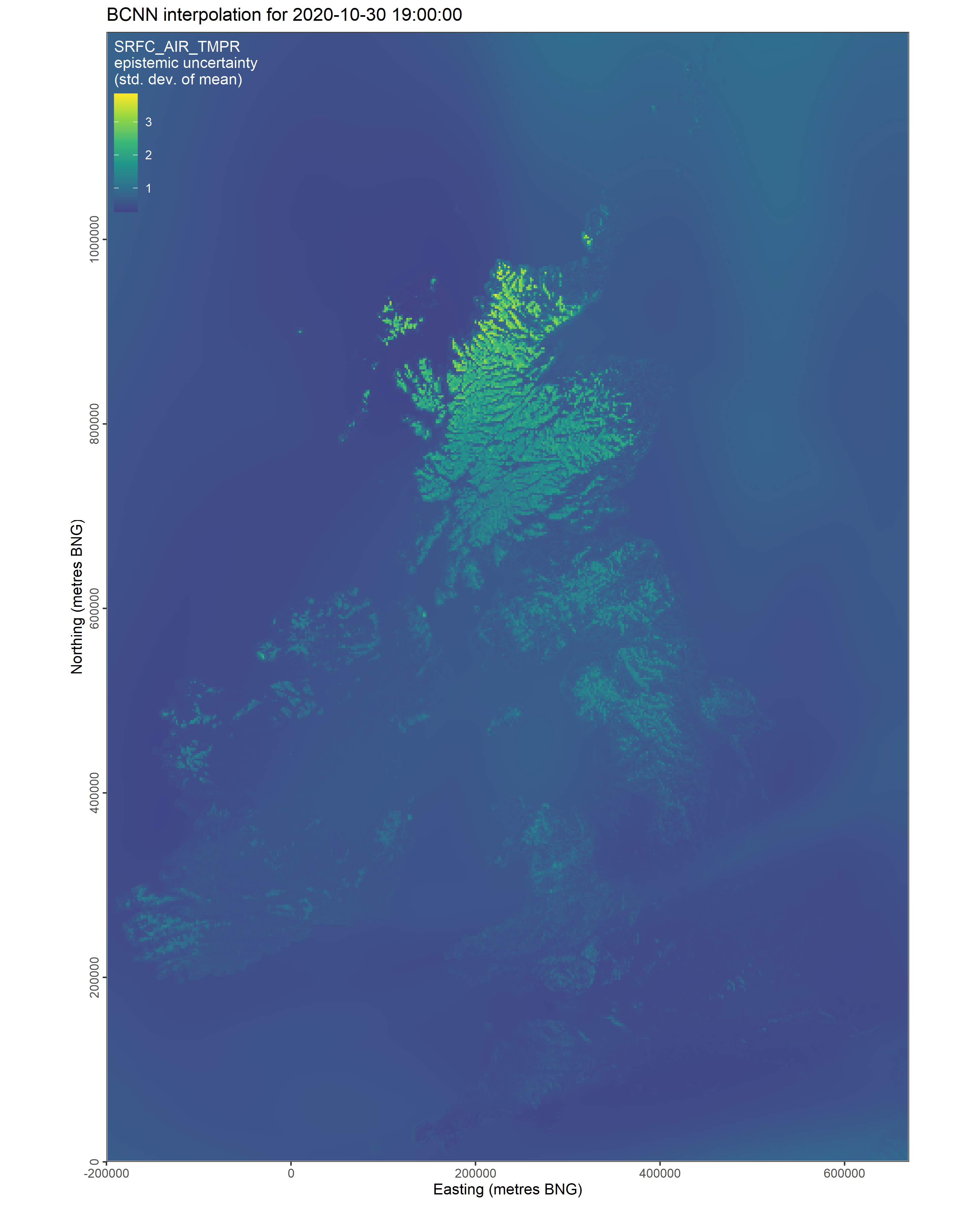}
    \caption{Epistemic uncertainty (standard deviation of the mean, °C) at the same snapshot in time.}
    \label{fig:epistemicmap}
\end{figure*}

\begin{figure*}[!htb]
    \centering
    \includegraphics[width=\textwidth]{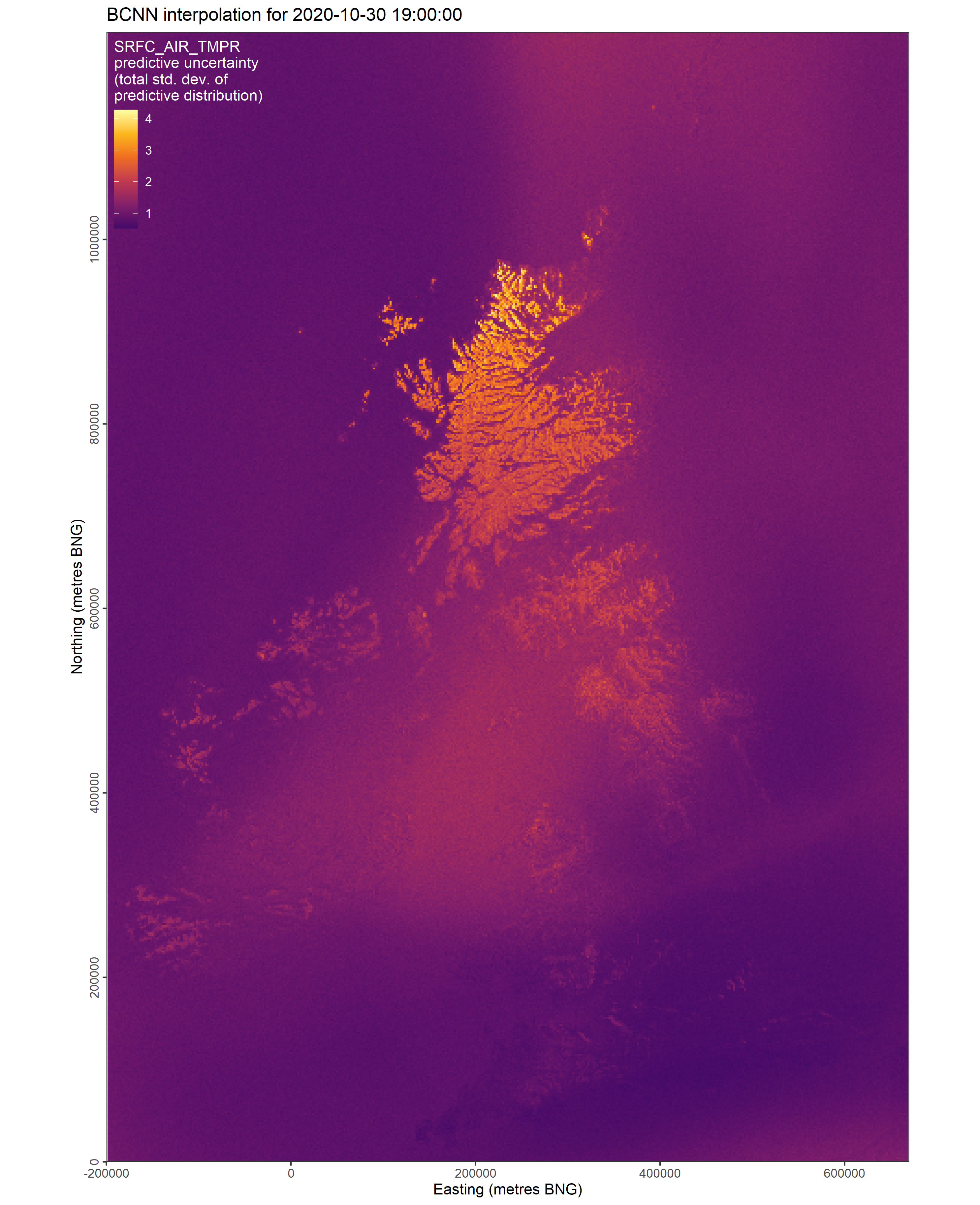}
    \caption{Total uncertainty (standard deviation, °C) of the predictive distribution at the same snapshot in time.}
    \label{fig:uncertaintymap}
\end{figure*}

Turning to the maps produced by our model, we can see that our deep learning approach produces detailed predictions which take account of surface topography. For any snapshot in time, we can obtain a map of the predicted mean (average value of $\mu_{s,t}$; \autoref{fig:meanmap}), the average aleatoric uncertainty (average value of $\sigma_{s,t}^2$; \autoref{fig:aleatoricmap}), the epistemic uncertainty of the mean (standard deviation of the posterior distribution of $\mu_{s,t}$; \autoref{fig:epistemicmap}), and the total uncertainty (standard deviation of the posterior predictive distribution; \autoref{fig:uncertaintymap}). Maps of any desired predictive quantiles, or other statistics of the posterior predictive distribution, can also be produced. In all such maps we can see that our deep learning approach produces predictions and predictive uncertainties that are highly spatially specific. In combination with the high quality of probabilistic calibration achieved (e.g. \autoref{fig:metrics} this indicates that our model is producing a predictive distribution that is both sharp and well-calibrated - the ideals for probabilistic predictions and forecasts as proposed by \citet{gneiting2007probabilistic}.

\begin{figure}[!htb]
    \centering
  \begin{subfigure}[t]{0.33\textwidth}
    \includegraphics[width=\linewidth, valign=t]{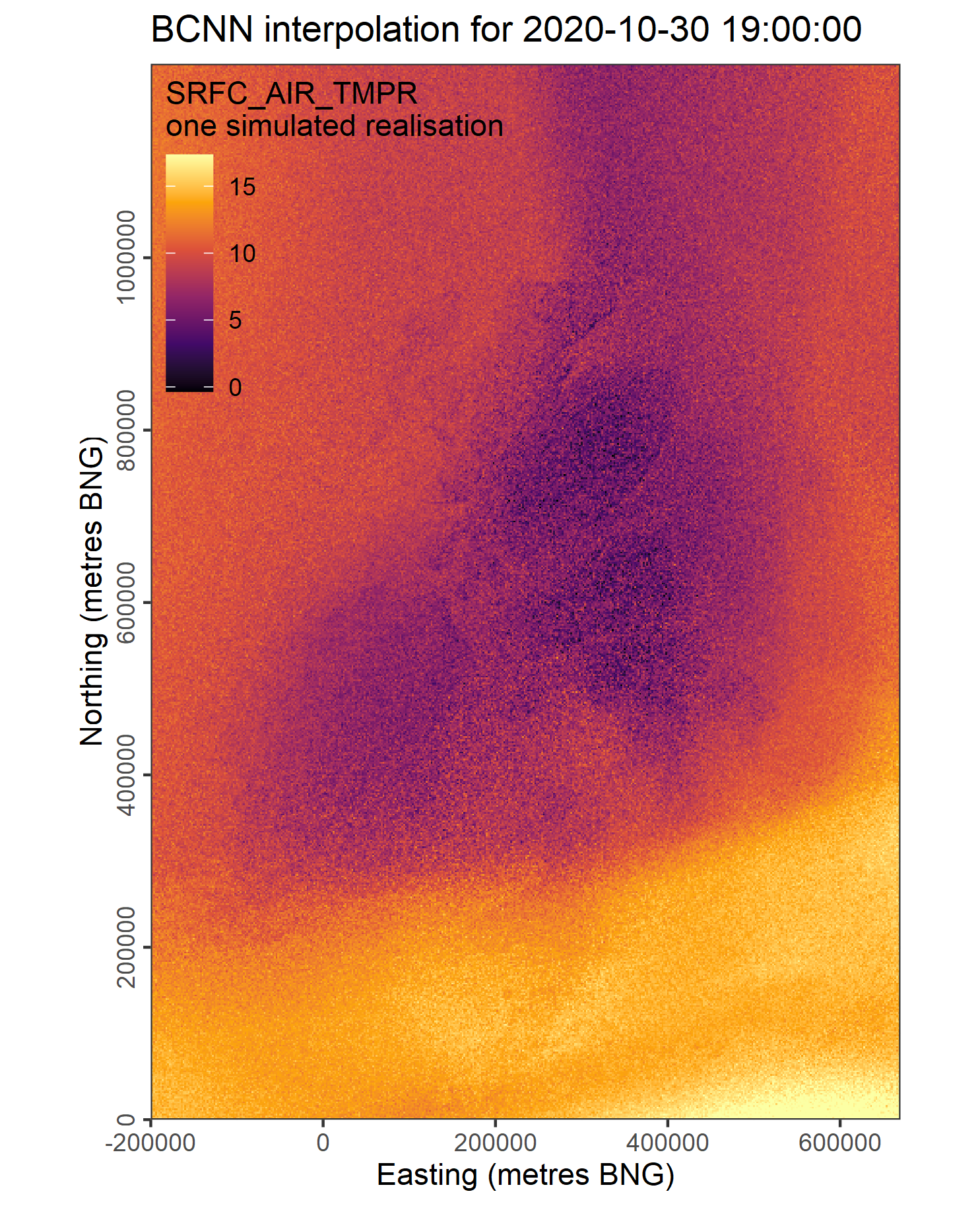}
    \includegraphics[width=\linewidth, valign=t]{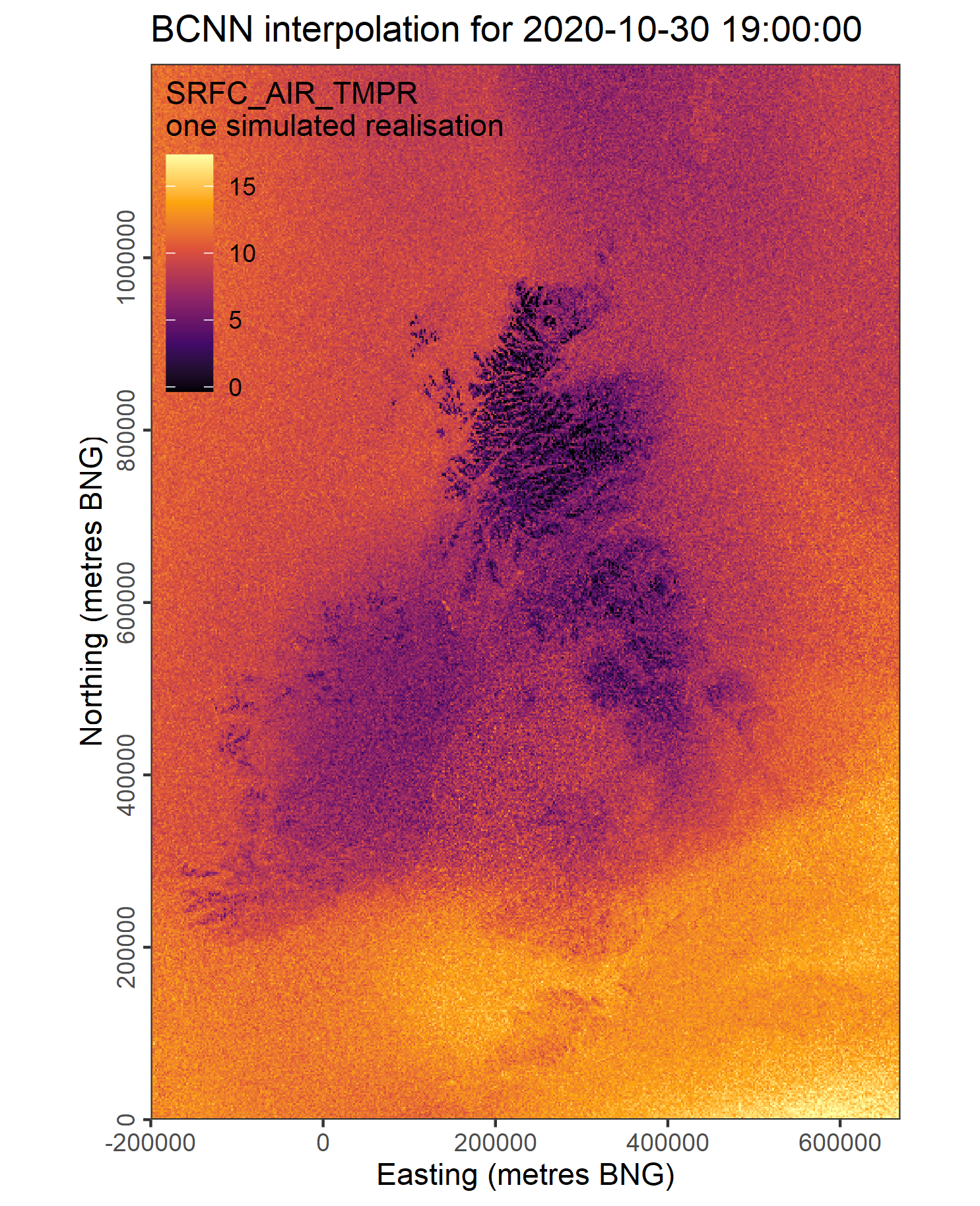}
  \end{subfigure}
  \begin{subfigure}[t]{0.33\textwidth}
    \includegraphics[width=\linewidth, valign=t]{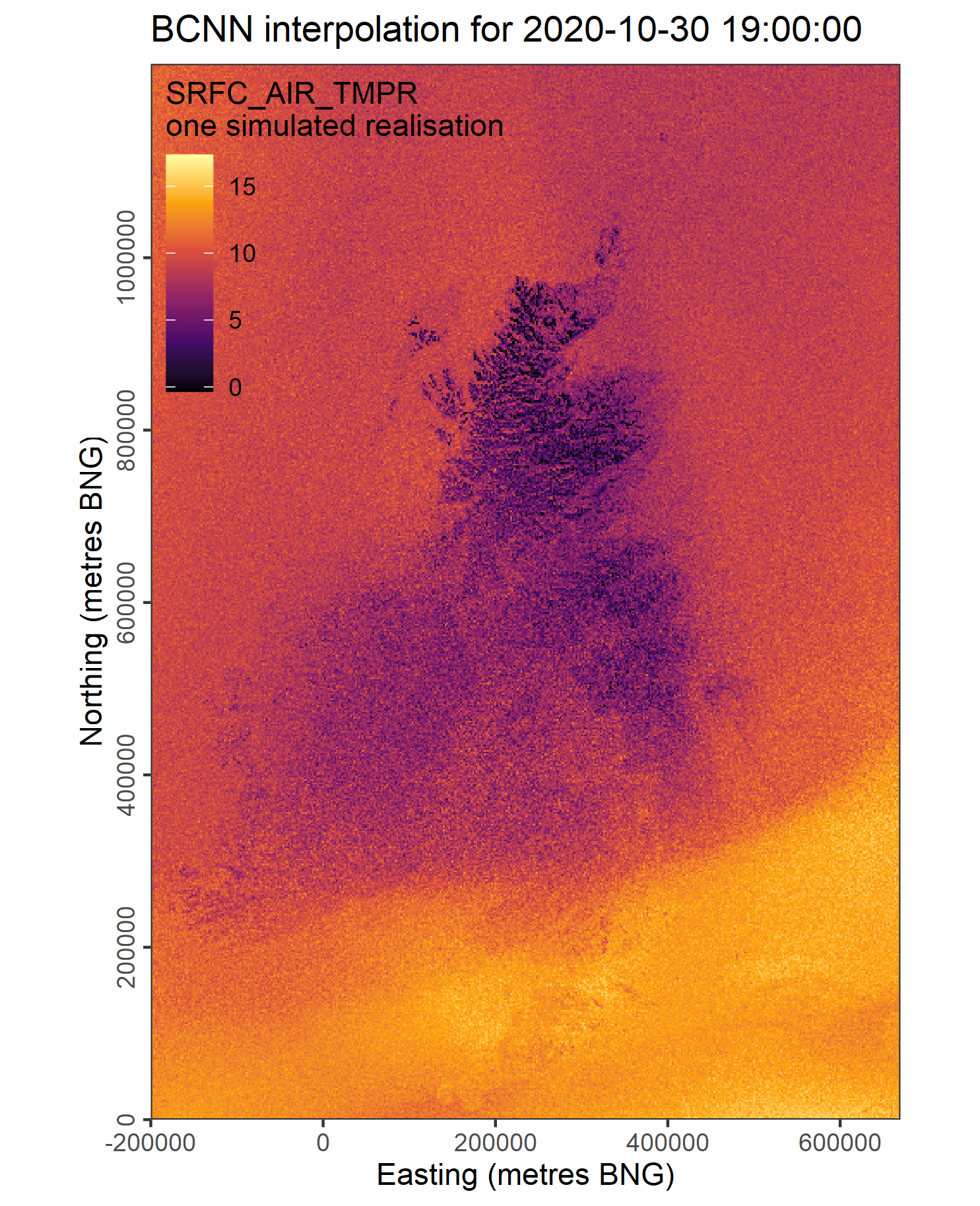}    \includegraphics[width=\linewidth, valign=t]{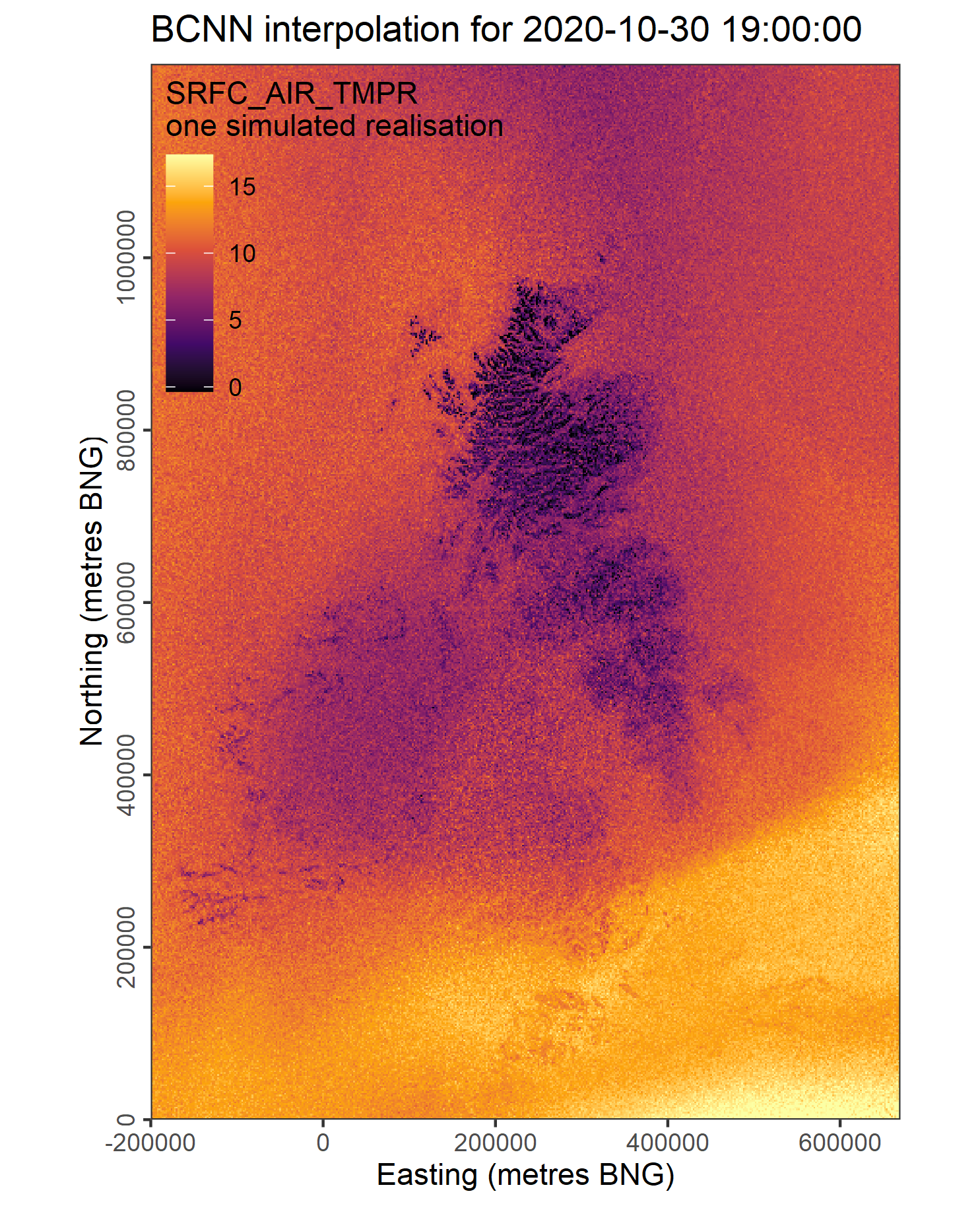}
  \end{subfigure}
  \begin{subfigure}[t]{0.33\textwidth}
    \includegraphics[width=\linewidth, valign=t]{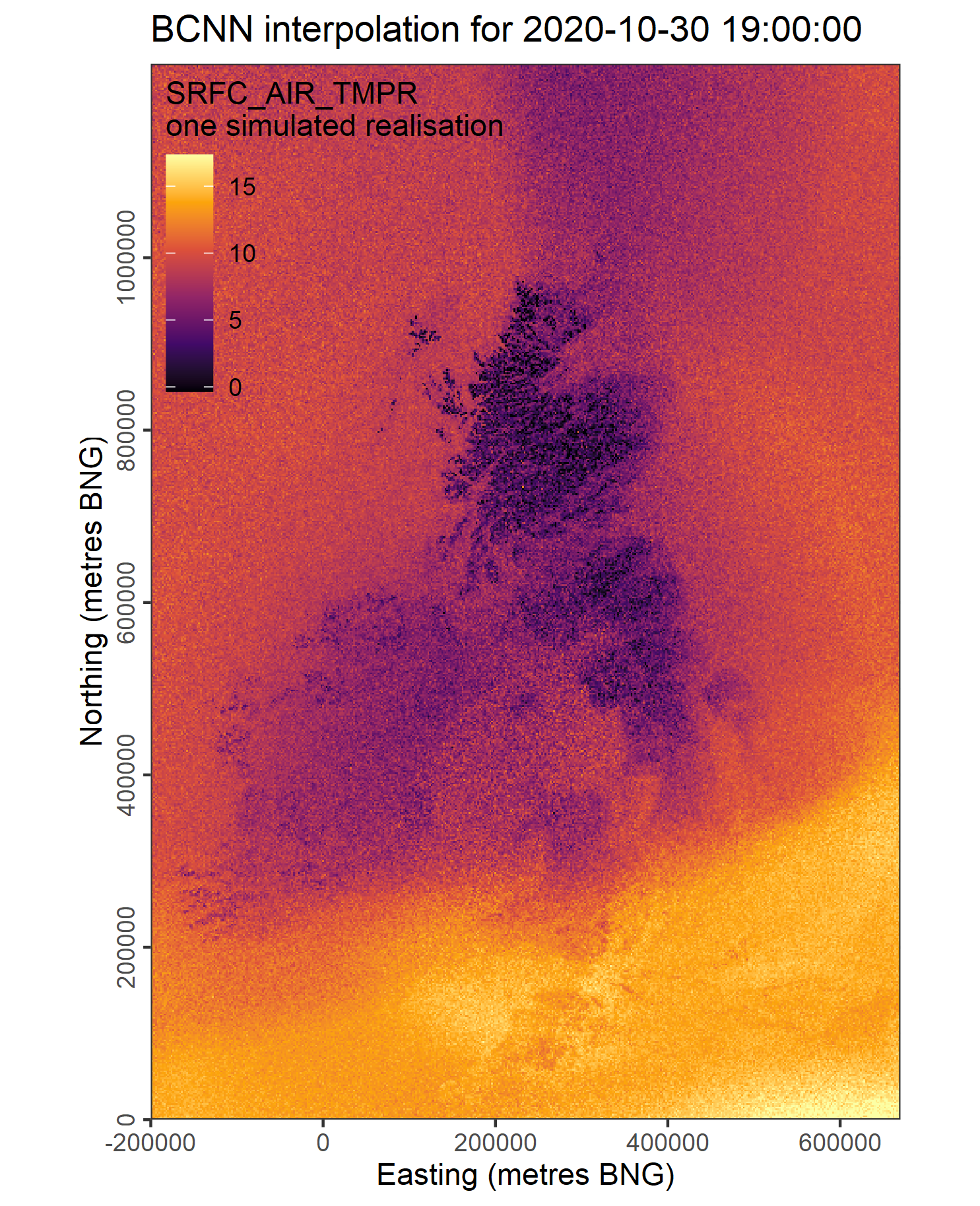}
    \includegraphics[width=\linewidth, valign=t]{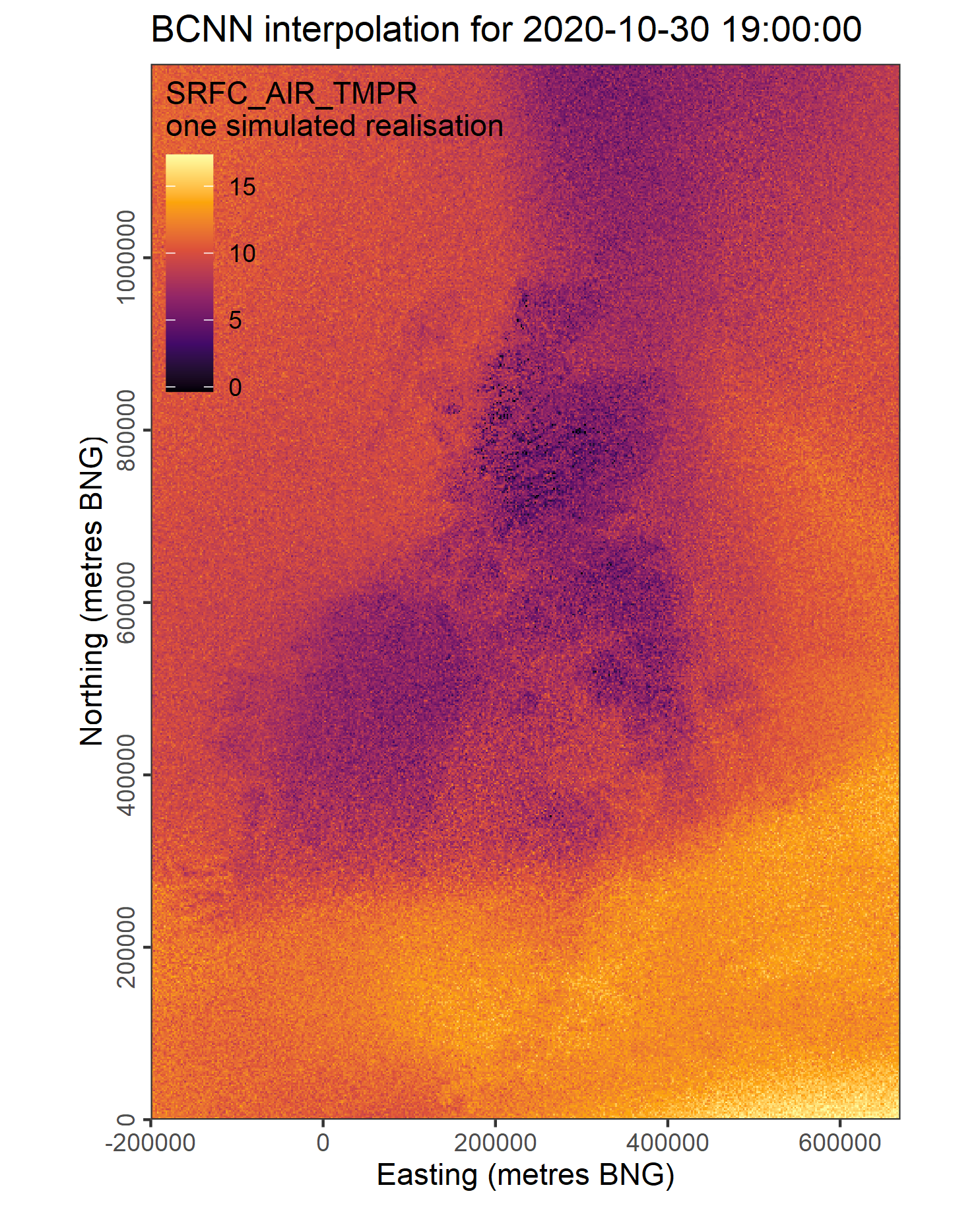}
  \end{subfigure}
      \caption{Six samples, or simulated realisations, from the posterior predictive distribution at 19:00 on 30/10/2020. As a collective (in the limit of infinite samples) such samples represent the total uncertainty in our spatio-temporal interpolation of surface air temperatures.}
    \label{fig:posteriormaps}
\end{figure}

\begin{figure}[!htb]
    \centering
  \begin{subfigure}[t]{0.33\textwidth}
    \includegraphics[width=\linewidth, valign=t]{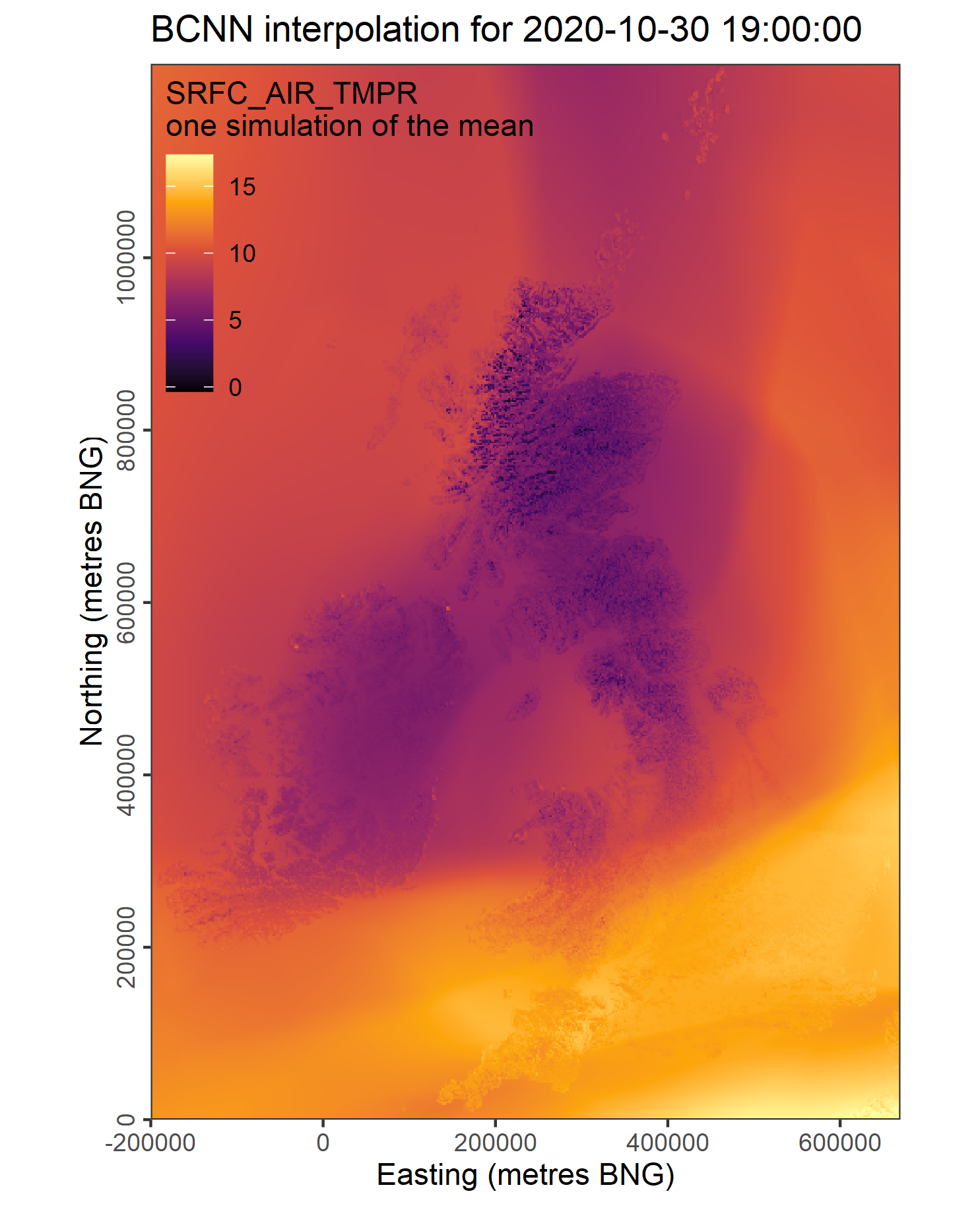}
    \includegraphics[width=\linewidth, valign=t]{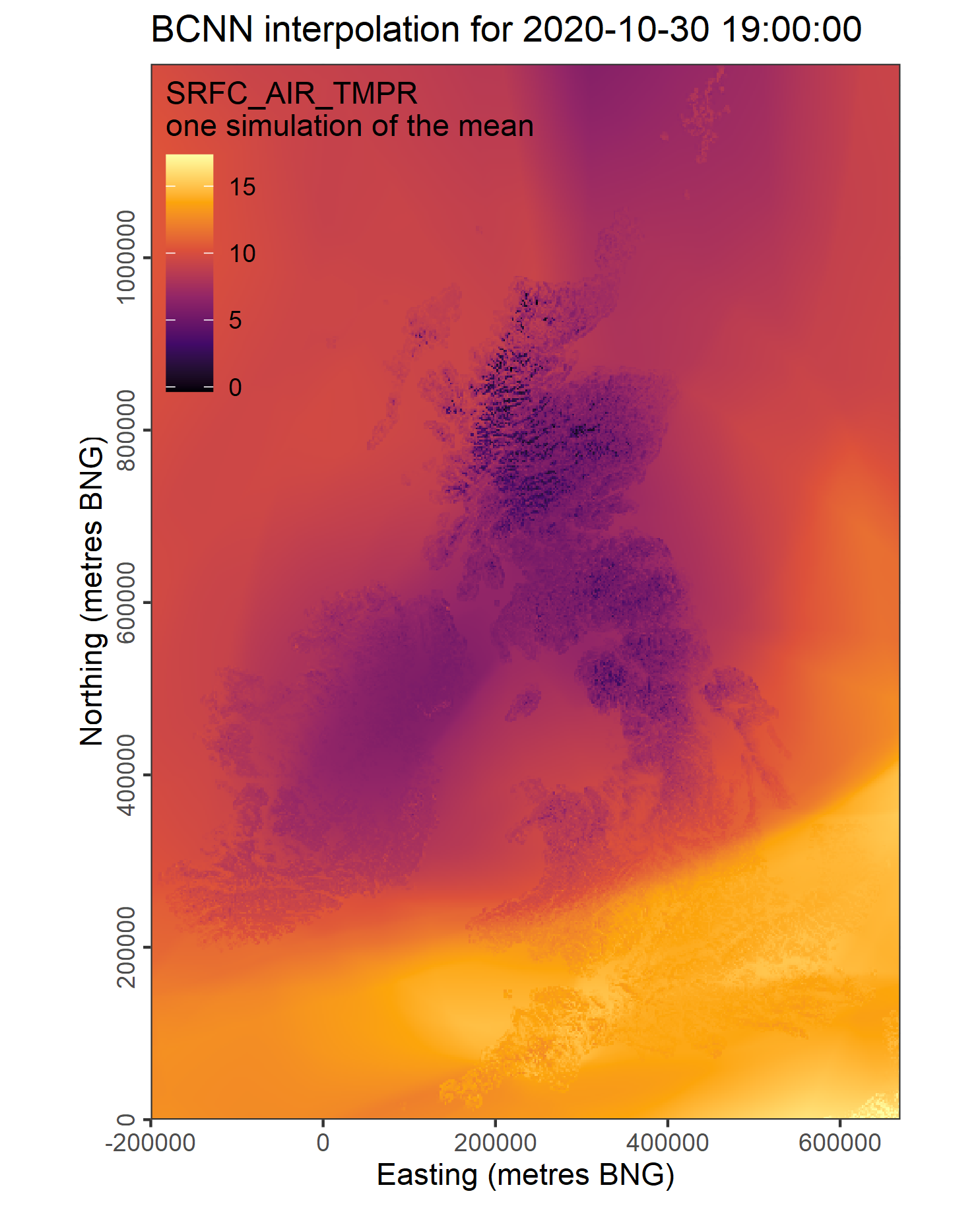}
  \end{subfigure}
  \begin{subfigure}[t]{0.33\textwidth}
    \includegraphics[width=\linewidth, valign=t]{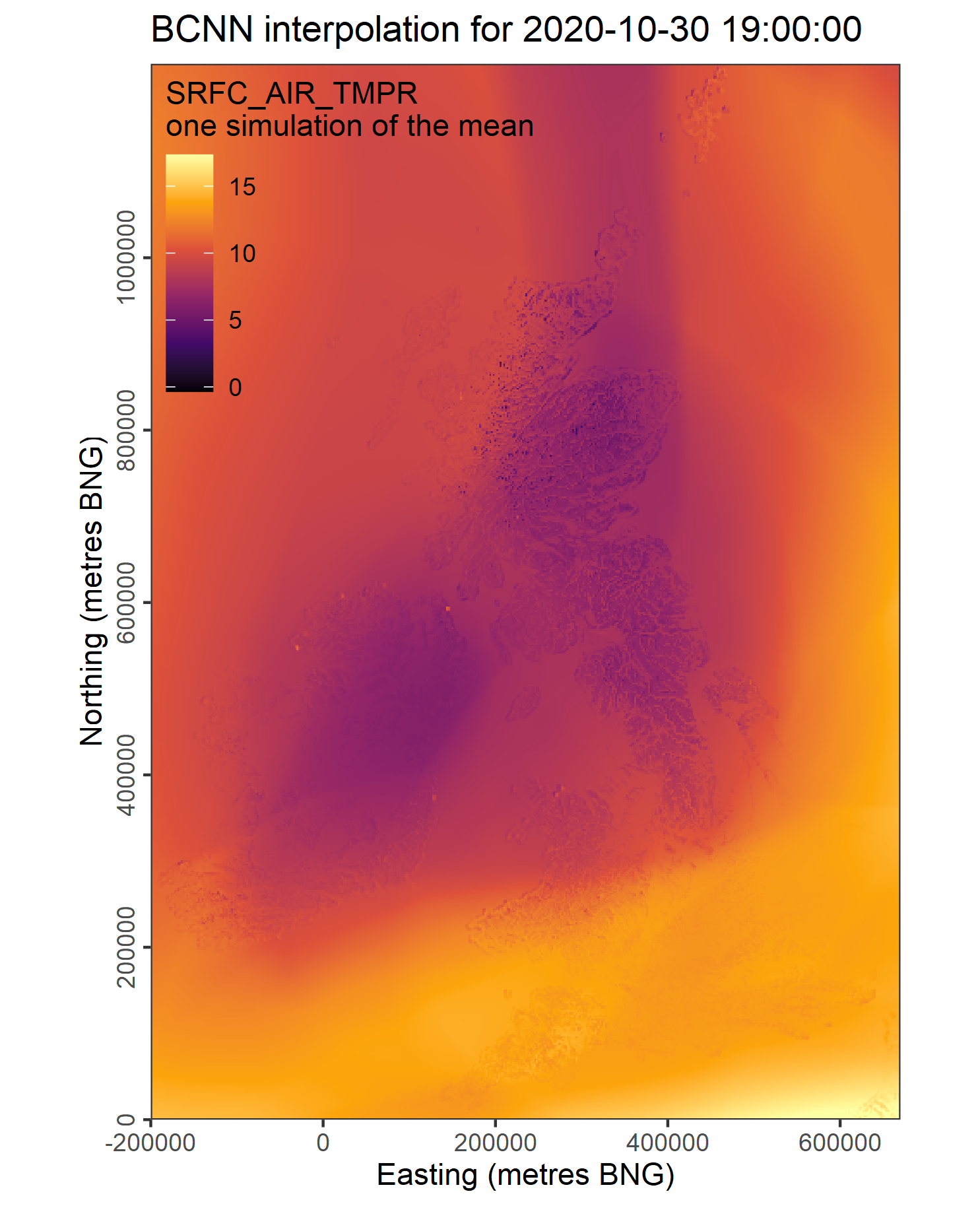}    
    \includegraphics[width=\linewidth, valign=t]{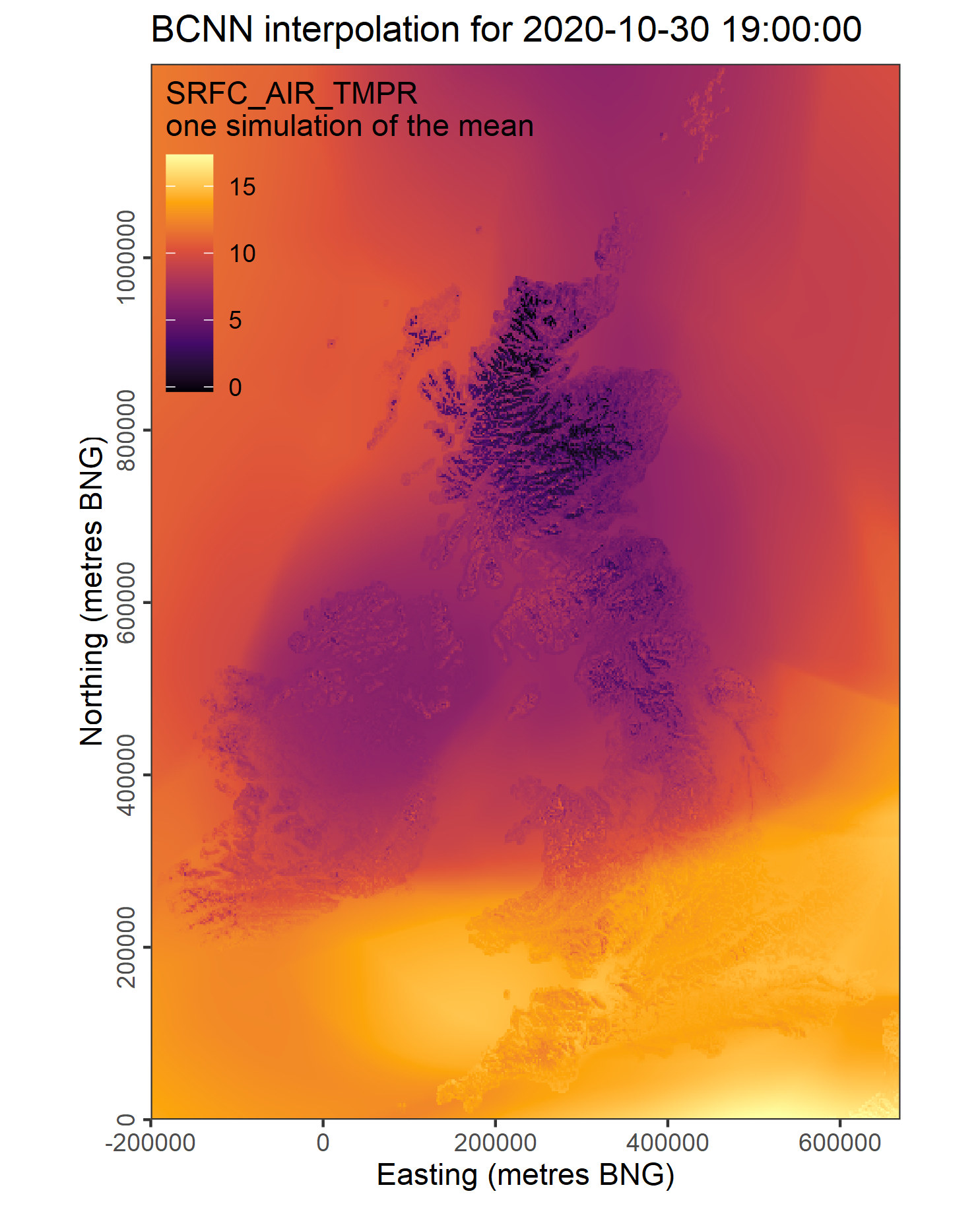}
  \end{subfigure}
  \begin{subfigure}[t]{0.33\textwidth}
    \includegraphics[width=\linewidth, valign=t]{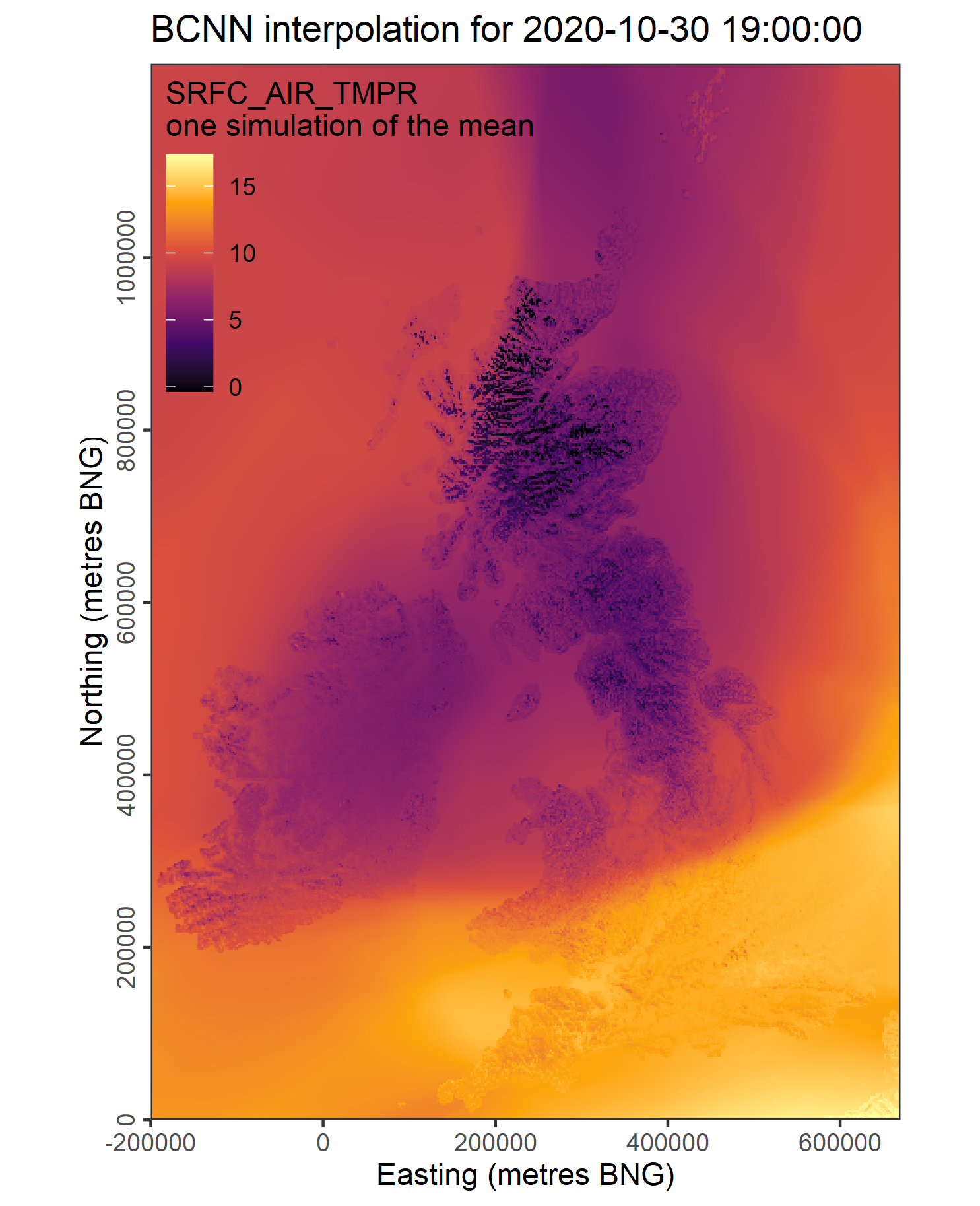}
    \includegraphics[width=\linewidth, valign=t]{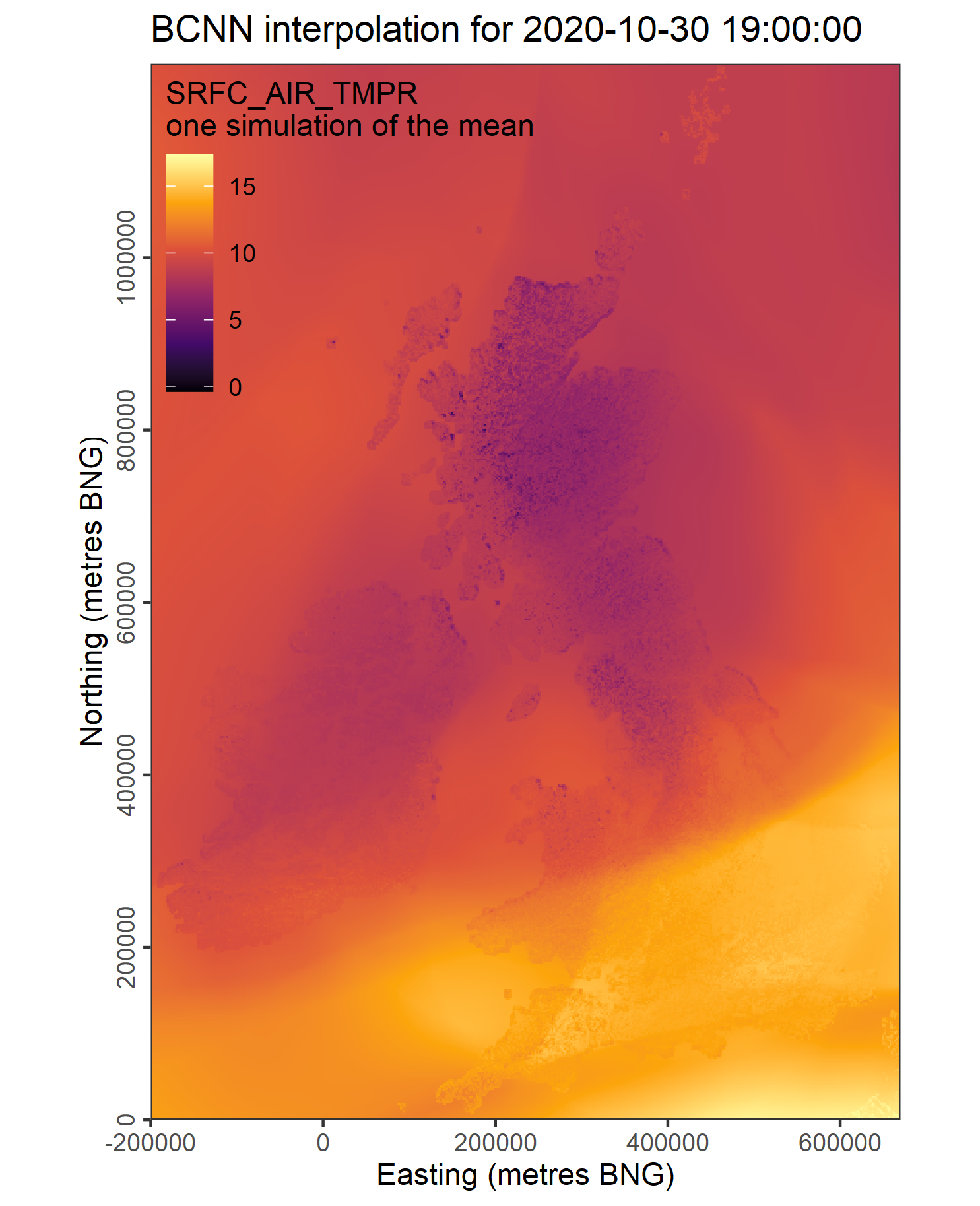}
  \end{subfigure}
      \caption{Six samples from the posterior distribution of the mean at 19:00 on 30/10/2020. As a collective (in the limit of infinite samples) such samples represent the epistemic uncertainty in our spatio-temporal interpolation of surface air temperatures.}
    \label{fig:posteriormeanmaps}
\end{figure}

Additionally, we can sample from the posterior distribution to generate simulated realisations of surface air temperature fields for any snapshot of time within the observed period. We can generate these both with the Gaussian output distribution active in order to achieve samples of the predictive distribution itself (including aleatoric uncertainty; \autoref{fig:posteriormaps}), or sample from only the posterior distribution of the mean (without independent noise from the Gaussian) in order to view alternative hypotheses for the mean temperature field at a given time (i.e. the epistemic uncertainty; \autoref{fig:posteriormeanmaps}). These simulated realisations help to convey the uncertainty in the model, by offering different explanations for plausible data generating processes.

\begin{figure}[!htb]
    \centering
  \begin{subfigure}[t]{1\textwidth}
    \includegraphics[width=\linewidth, valign=t]{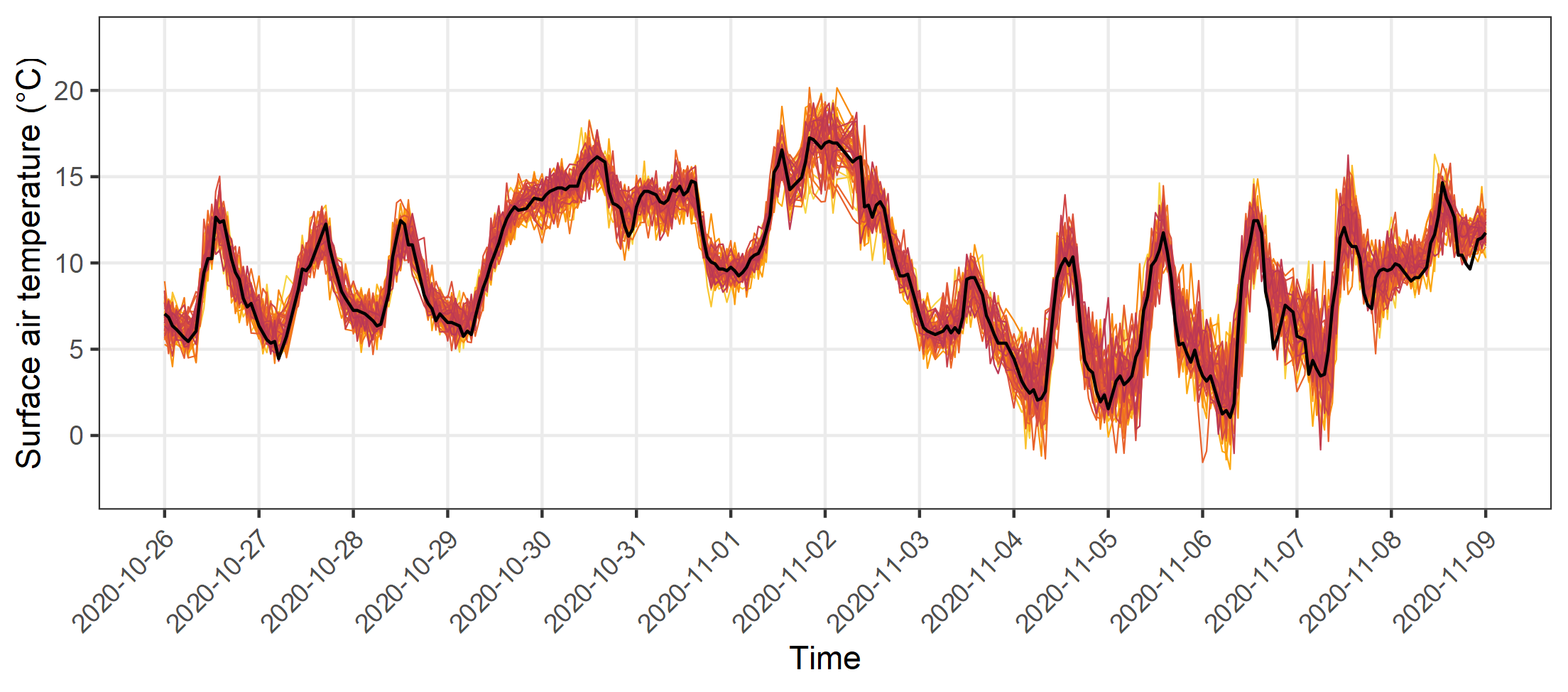}
    \includegraphics[width=\linewidth, valign=t]{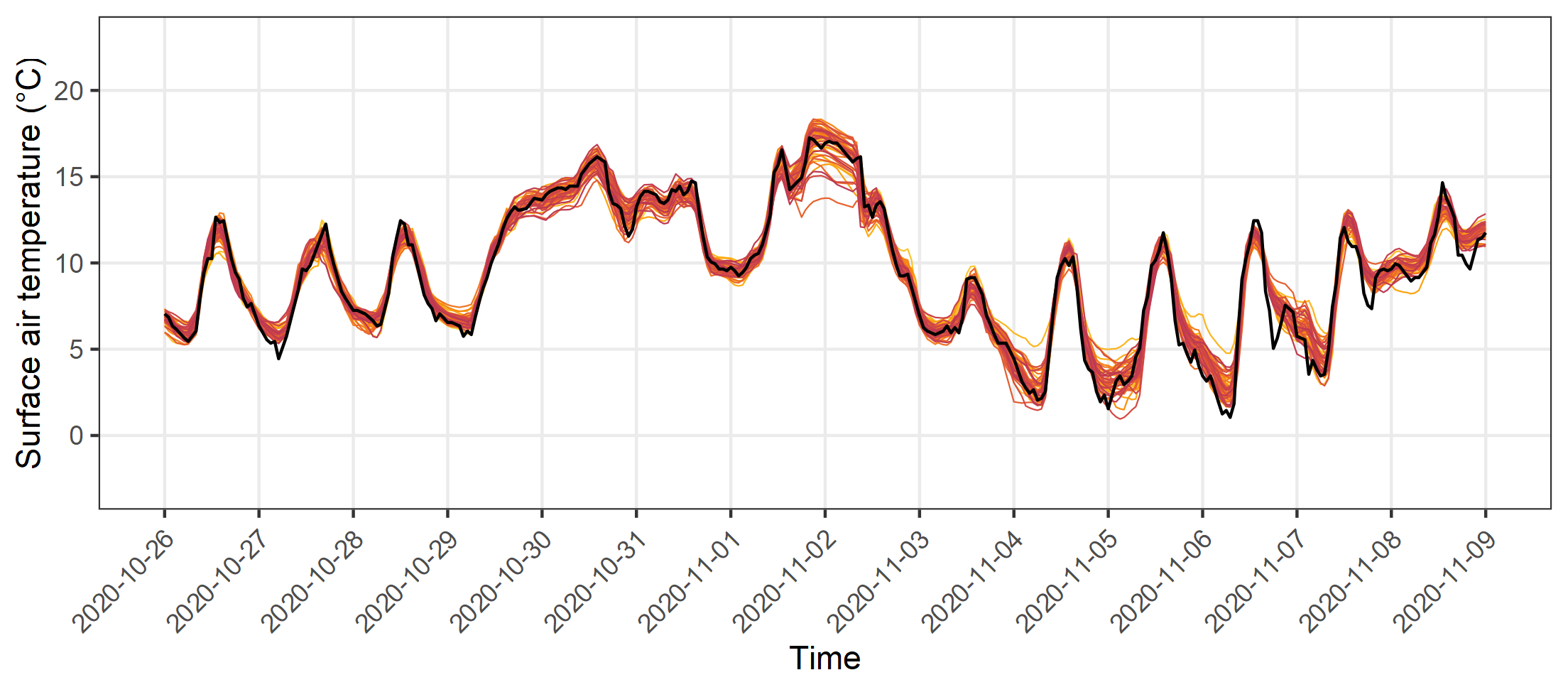}
  \end{subfigure}
      \caption{Time series of samples generated from the trained model, showing total uncertainty (top, epistemic and aleatoric uncertainty), and uncertainty in just the predicted mean (bottom, epistemic uncertainty only) for predictions on a held-out test site.}
    \label{fig:posteriortimeseries}
\end{figure}

To visualise the output of the model through time, rather than purely in space, we can compare samples from the model (again with and without aleatoric uncertainty included) to observations recorded at a held out test site as timeseries (\autoref{fig:posteriortimeseries}). As is indicated by the overall model fit and calibration metrics (\autoref{fig:metrics}, the predictive performance for held out test sites is good - we can see in the timeseries of samples from the model that samples of the mean track the observations quite closely (but do not track noise in the observations) meanwhile, samples from the posterior predictive distribution, with aleatoric uncertainty included, do a good job of covering the distribution of observations, including noise. Both epistemic and aleatoric uncertainty vary through time (and space, as we saw in \autoref{fig:aleatoricmap} and \autoref{fig:epistemicmap}). Animations of the model output (perhaps the best way to view spatio-temporal model output) are available to view at https://github.com/charliekirkwood/animations.

The role of the model we present here, as a spatio-temporal interpolator of weather data, is similar to the role that would traditionally be filled by numerical hindcasting \citep[e.g.][]{palmer2004development}. This is where the same physics-based numerical weather prediction models used for forecasting are fitted retrospectively to historic weather observations, to provide a `best fit' of historic weather conditions. In order to provide an indication of uncertainty, ensemble hindcasts can also be run, but it is generally the case that numerical weather prediction ensembles are underdispersive in relation to observations \citep[e.g.][]{gneiting2005calibrated}. By providing well-calibrated spatio-temporal interpolations, our deep learning approach may have the potential to provide a probabilistically-superior (and computationally cheaper) alternative to numerical hindcasting, despite our model having no notion of the physical equations that govern atmospheric dynamics \citep[i.e. Navier-Stokes;][]{kimura2002numerical}. The level of detail of spatial structure captured by the model will be limited by a combination of the resolution of auxiliary information (gridded terrain elevation data in this case) and the spatial density of observations, but the model remains free to provide predictions for any point in space. The resolution, or spatial precision, of our proposed approach can naturally improve as the density of observations, and the resolution of auxiliary information, increases.

It is interesting to observe the difference between samples of our model's posterior distribution both with and without aleatoric uncertainty --- the independent noise provided by the Gaussian output distribution --- included (e.g. by comparing the top and bottom of \autoref{fig:posteriortimeseries}, or comparing \autoref{fig:posteriormaps} with \autoref{fig:posteriormeanmaps}). As can be seen in \autoref{fig:posteriortimeseries}, the independent noise of our Gaussian output distribution is required in order to provide well-calibrated coverage in relation to observations (at least in our setup, in which independent noise is a part of the model). Without this aleatoric uncertainty included, the distribution over our plausible mean functions would be underdispersive in relation to the observations. This has parallels in the setup of numerical weather prediction and hindcasting \citep[e.g.][]{rawlins2007met, rougier2013model, bauer2015quiet}, in which ensembles tend to be underdispersive in part for the same reason: that while these numerical ensemble members do capture epistemic uncertainty in initial conditions \cite{rougier2013intractable} and perhaps across model parameters \cite{leutbecher2008ensemble}, they tend not to model aleatoric uncertainty. In order to achieve well-calibrated numerical weather forecasts, statistical-post processing must therefore be used, such as Bayesian model averaging in which individual ensemble members are `dressed' with suitably scaled Gaussian noise \citep{raftery2005using}, thus effectively transforming the underdispersed ensemble at the bottom of \autoref{fig:posteriortimeseries} to the well-calibrated ensemble at the top of \autoref{fig:posteriortimeseries}. It is perhaps another strength of our Bayesian deep learning approach that `ensemble' predictions of both forms (with and without aleatoric uncertainty) can be generated equally easily by sampling from the same model, and that our full posterior predictive distribution (which includes aleatoric uncertainty) is innately well-calibrated and requires no subsequent post-processing.

\clearpage

\clearpage

\clearpage

\clearpage

\clearpage

\clearpage

\clearpage

\section*{Conclusions}

We have presented a deep learning approach that provides well-calibrated outlier-corrected spatio-temporal interpolation of crowd-sourced weather observations. Our deep mixture density network approach to outlier classification unifies outlier detection and correction as part of a same single probabilistic data modelling process, which provides a more streamlined modelling and model-checking workflow compared to alternative two stage techniques (in which outlier detection and filtering is performed separately prior to data modelling). 

Our unified approach allows us to, through a single probabilistic data model (our Bayesian deep neural network), generate high fidelity spatio-temporal predictions from historic crowd-sourced weather observations. The ultimate functionality is therefore similar to that of numerical hindcasting or reanalysis, but our approach is likely to be computationally cheaper and our predictions are innately well-calibrated, requiring no post-processing. By providing a full predictive distribution, the uncertainty of predictions is fully quantified, therefore making our output useful to decision makers. The predictive uncertainty can also be viewed and mapped as its two separate components: aleatoric uncertainty, or irreducible uncertainty in the data, and epistemic uncertainty, or reducible uncertainty about our state of knowledge. In addition our predictions can be provided at any point in space and time, therefore catering for hyper-local scales, and so may be viewed as satisfying the requirements of a `models of everywhere' approach to harnessing Internet of Things (IOT) type weather observations. On the basis of all these benefits we therefore consider our approach to have potentially powerful applications for quality control, data assimilation and climatological studies that maximise the utility of IOT data for environmental modelling applications in an increasingly data-rich world.

\subsection*{Acknowledgements}
We acknowledge funding from the UK's Engineering and Physical Sciences Research Council (EPSRC project ref: 2071900) and from the UK Met Office, by which CK's PhD studentship is funded.

\subsection*{Data availability}

The code to reproduce this study is available at https://github.com/charliekirkwood/wowpaper and includes functions to download NASA's SRTM elevation data via the raster package in R. Data from the Met Office's Weather Observation Website can be downloaded from https://wow.metoffice.gov.uk/

\bibliographystyle{unsrtnat}  
\bibliography{ref}

\end{document}